%% file: acl_latex.tex
\pdfoutput=1

\documentclass[11pt]{article}

\usepackage[final]{acl}

\usepackage{times}
\usepackage{latexsym}
\usepackage{array}
\newcolumntype{H}{>{\setbox0=\hbox\bgroup}c<{\egroup}@{}}
\usepackage[T1]{fontenc}

\usepackage[utf8]{inputenc}

\usepackage{microtype}

\usepackage{inconsolata}

\usepackage{url}
\usepackage{twemojis}
\usepackage{dsfont}
\usepackage{bm}
\usepackage{bbm}
\usepackage{graphicx}
\usepackage{color}
\usepackage{multicol}
\usepackage{multirow}
\usepackage{wrapfig,lipsum,booktabs}
\usepackage[ruled,vlined,linesnumbered]{algorithm2e}
\usepackage{pgfplots}
\pgfplotsset{compat=1.12}
\usepackage{xcolor}
\usepackage{tikz}
\usepackage{xspace}
\usepackage{makecell}
\usepackage{soul}
\usepackage{amsmath,amsfonts,amssymb}
\usepackage{subcaption}
\usetikzlibrary{calc}
\usepgfplotslibrary{groupplots}
\usetikzlibrary{angles,quotes} 
\usetikzlibrary{shapes,arrows}
\usetikzlibrary{backgrounds}
\usetikzlibrary{matrix}
\usepackage{tikz-3dplot}
\usepackage{hyperref}
\usepackage{cleveref}
\usepackage{paralist}
\usepackage{cancel}
\usepackage{xspace}
\usepackage{todonotes}
\usepackage{tabu}
\usepackage{rotating}
\usepackage{etoolbox}
\usepackage{adjustbox}
\usepackage{enumerate}
\usepackage{enumitem}
\setitemize{noitemsep,topsep=0pt,parsep=0pt,partopsep=0pt}
\setenumerate{noitemsep,topsep=0pt,parsep=0pt,partopsep=0pt}
\usepackage{pifont}
\usepackage{cancel}
\usepackage{lipsum}
\usepackage{listings,lstautogobble}
\usepackage{fancyvrb}
\usepackage{fvextra}
\usepackage{caption}
\usepackage{pgf-pie} 
\usepackage{array, makecell}
\usepackage{wrapfig}

\title{Beyond Probabilities: Unveiling the Misalignment in Evaluating Large Language Models}

\author{
Chenyang Lyu\textsuperscript{1$\dagger$}\quad Minghao Wu\textsuperscript{2$\dagger$}\quad Alham Fikri Aji\textsuperscript{1} \\
\textsuperscript{1}Mohamed bin Zayed University of Artificial Intelligence\\
\textsuperscript{2}Monash University\\
\texttt{\{chenyang.lyu,alham.fikri\}@mbzuai.ac.ae}\quad\texttt{minghao.wu@monash.edu}
}

\begin{document}
\maketitle
\def\thefootnote{$\dagger$}\footnotetext{Equal contribution}
\begin{abstract}

Large Language Models (LLMs) have demonstrated remarkable capabilities across various applications, fundamentally reshaping the landscape of natural language processing (NLP) research. However, recent evaluation frameworks often rely on the output probabilities of LLMs for predictions, primarily due to computational constraints, diverging from real-world LLM usage scenarios. While widely employed, the efficacy of these probability-based evaluation strategies remains an open research question. This study aims to scrutinize the validity of such probability-based evaluation methods within the context of using LLMs for Multiple Choice Questions~(MCQs), highlighting their inherent limitations. Our empirical investigation reveals that the prevalent probability-based evaluation method inadequately aligns with generation-based prediction. Furthermore, current evaluation frameworks typically assess LLMs through predictive tasks based on output probabilities rather than directly generating responses, owing to computational limitations. We illustrate that these probability-based approaches do not effectively correspond with generative predictions. The outcomes of our study can enhance the understanding of LLM evaluation methodologies and provide insights for future research in this domain.  
\end{abstract}

\input{1_introduction}
\input{2_background}
\input{4_results}

\input{5_discussion}
\input{6_related_work}
\input{7_conclusion}
\input{8_limitations}

\bibliography{anthology,custom}

\input{9_appendix}

\end{document}

%% file: 1_introduction.tex
\section{Introduction}

\begin{figure*}[t]
    \centering
    \includegraphics[width=\linewidth]{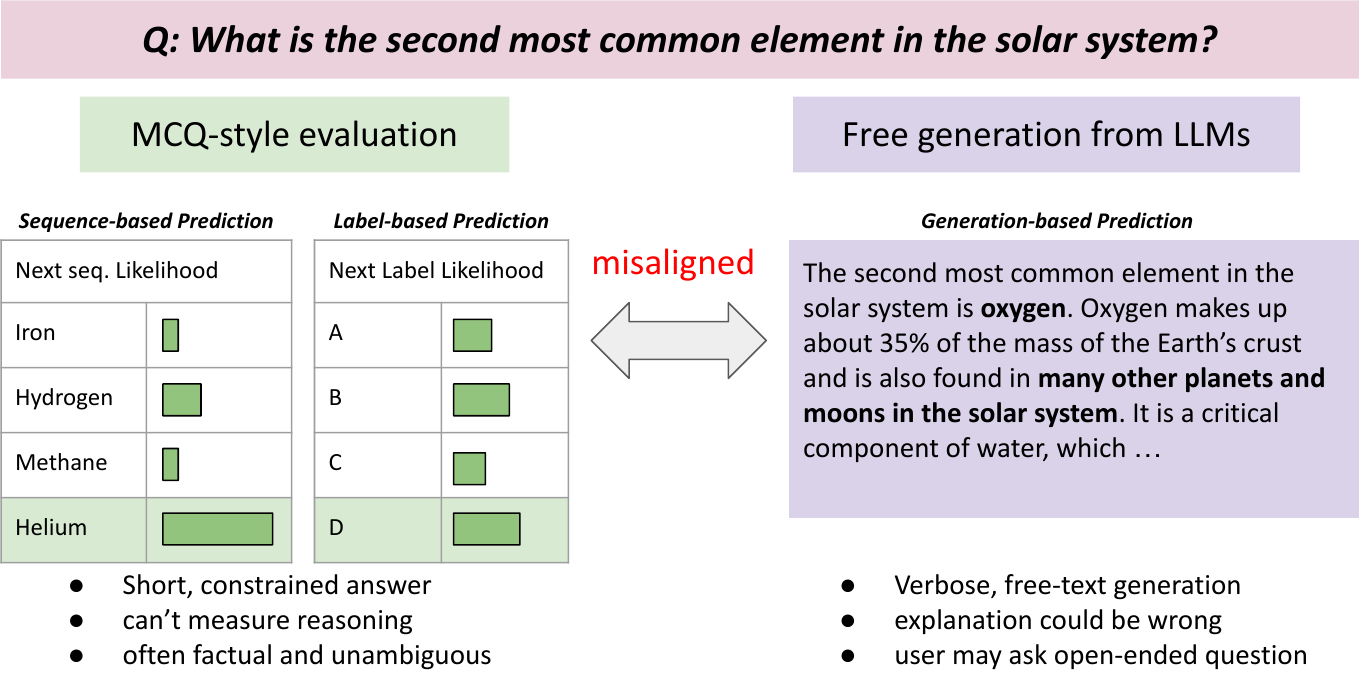}
    \caption{An illustration of label-based, sequence-based and generation-based predictions for evaluating LLMs on NLP benchmarks.}
    \label{fig:intro}
\end{figure*}

Large Language Models (LLMs) have significantly advanced the field of natural language processing (NLP), reshaping the paradigms in NLP research and application \citep{ouyang2022training, wei2022finetuned, DBLP:conf/iclr/SanhWRBSACSRDBX22, DBLP:journals/corr/abs-2210-11416, DBLP:journals/corr/abs-2303-08774, DBLP:journals/corr/abs-2305-10403, DBLP:journals/corr/abs-2302-13971,touvron2023llama2,jiang2023mistral}. As the scale of model parameters of language models expands from the million to billion or even trillion levels, a proficient LLM is expected to exhibit a broad mastery across various tasks. Recent works aim to assess LLMs comprehensively by aggregating a substantial array of NLP benchmarks \citep{DBLP:journals/corr/abs-2206-04615, DBLP:conf/iclr/SanhWRBSACSRDBX22, DBLP:journals/corr/abs-2211-09110, DBLP:journals/corr/abs-2301-13688}. Additionally, there exists a line of research that curates human exam questions to challenge LLMs \citep{DBLP:conf/iclr/HendrycksBBZMSS21_mmlu, DBLP:journals/corr/abs-2305-08322, DBLP:journals/corr/abs-2306-09212,koto-etal-2023-large}. The collected questions and NLP benchmarks are adapted into prompts via standardized templates.

Due to computational constraints, recent evaluation frameworks commonly adopt the approach of selecting the option with the highest probability as the prediction of LLMs, as illustrated in \autoref{fig:intro}. These frameworks employ either \textit{label-based prediction}, which assesses the probability of the next token output, or \textit{sequence-based prediction}, which evaluates the probability of an entire option, ultimately selecting the option with the highest probability as the LLM's prediction. However, these probability-based evaluation methodologies introduce a misalignment between evaluation procedures and real-world application scenarios, where LLMs are typically tasked with generating responses to user queries. This misalignment raises an important question: \textit{Is the probability-based evaluation method sufficient to accurately assess the capabilities of LLMs?}



In this position study, we argue that the current LLM evaluation and leaderboard misalign the actual LLM capabilities. We examine three prediction methodologies: generation-based, label-based, and sequence-based predictions. We conducted extensive experiments across LLMs with varying model sizes on three prominent benchmarks: MMLU \citep{DBLP:conf/iclr/HendrycksBBZMSS21_mmlu}, TruthfulQA \citep{lin-etal-2022-truthfulqa}, and Belebele \citep{bandarkar2023belebele}. Our findings reveal a significant disconnect between probability-based methods and generation-based predictions. Even when predictions are correct, the consistency between probability-based methods and generation-based predictions remains notably low. We additionally find that many of these multiple-choice NLP benchmark rankings do not agree with human preference for free-text generation output.
Consequently, these results raise serious doubts about the reliability of evaluation outcomes derived from popular benchmarks reliant on probability-based methods. In conclusion, our research emphasizes the urgent need for an evaluation approach that ensures accurate and reliable assessments of LLM capabilities, more closely aligned with real-world usage scenarios. In next section, we will discuss the course of the development and paradigm of the evaluation of LLMs.

%% file: 2_background.tex
\section{Evaluating Large Language Models}

\subsection{Challenges in Evaluating Large Language Models}
\label{sec:challenge}

The advancement of LLMs has substantially broadened their capabilities, transcending conventional NLP tasks. They now demonstrate proficiency in tackling intricate prompts and a wide spectrum of open-ended inquiries. However, unlike tasks with definitive solutions, open-ended questions lack a single correct answer, making it difficult to gauge the LLM's performance.

Recently, human evaluators have been deployed to appraise responses to open-ended questions using two primary methods. Firstly, evaluators assign scores based on specific criteria such as accuracy and relevance \citep{wang-etal-2023-self-instruct, DBLP:journals/corr/abs-2305-11206}. Alternatively, they conduct comparative assessments by selecting the preferred answer among two distinct LLM responses to the same question \citep{DBLP:journals/corr/abs-2112-00861, DBLP:journals/corr/abs-2204-05862, DBLP:journals/corr/abs-2306-05685}. However, manual evaluation faces significant scalability challenges due to the high costs associated with human judges. Moreover, recent studies indicate that human evaluators often favor longer and more fluent responses, even if they contain factual inaccuracies \citep{DBLP:journals/corr/abs-2307-03025}. Additionally, ensuring the trustworthiness of evaluations presents a concern, as crowd-annotators increasingly rely on tools like LLMs for assistance \citep{DBLP:journals/corr/abs-2306-07899}, raising questions about the purely human-based nature of evaluations. Moreover, maintaining consistent evaluation quality across a large team of evaluators necessitates extensive coordination and rigorous standardization. Recent research highlights low consistency among human evaluators when assessing LLM responses to open-ended questions.

Another approach to evaluating generative LLMs involves utilizing a stronger LLM as the evaluator, offering greater scalability compared to human judges \citep{DBLP:journals/corr/abs-2306-05685, DBLP:journals/corr/abs-2307-03025, liu-etal-2023-g}. However, LLM judges may exhibit biases in their assessments, influenced by factors such as the order and length of answers, as well as their fluency. Furthermore, commonly used LLM judges, like GPT-4~\cite{openai2023gpt4}, often operate on public yet black-box systems, posing challenges in ensuring the reproducibility and transparency of the evaluation process.

\subsection{Multiple Choice Question as a Proxy}
\label{sec:proxy}

Due to the challenges discussed in \autoref{sec:challenge}, recent works commonly convert the multiple-choice questions (MCQs) in human exams to prompts using standard template. The responses generated by the LLMs are then compared against the human-crafted ground truth, allowing for an assessment of the model's accuracy.  This process streamlines the evaluation and provides a clear metric for understanding the capabilities of LLMs.

Recent frameworks frequently utilize the output probabilities from LLMs across various options for making predictions, to ensure that the prediction from the LLM is among these options, given the unpredictability of the text generated by LLMs. For example, as illustrated in \autoref{fig:intro}, when presented with the question and the candidate choices, some approaches compare the probabilities predicted by the model based solely on the option letters \citep{DBLP:conf/iclr/HendrycksBBZMSS21_mmlu},\footnote{\url{https://github.com/hendrycks/test}} while others consider the probability of each token and aggregate them \citep{eval-harness}.\footnote{\url{https://github.com/EleutherAI/lm-evaluation-harness}} 


\subsection{Misalignment between MCQ and User-Facing Interaction}

We argue that MCQ-proxy might not always reflect the actual performance of LLM under user-facing free-text generation. In MCQ, LLM output is restricted to a limited set of answers; hence, their answer might be different under unrestricted generation. MCQ benchmarks also often only look for a short and direct answer, whereas user-facing interaction expects the LLM to provide a verbose answer; especially after preference tuning. Hence, MCQ benchmarks are not suitable for measuring the nuanced answers of LLMs.

Additionally, prior studies have shown LLM's brittleness under MCQ benchmarks, e.g., on how the option order is presented~\cite{zheng2023large, pezeshkpour2023large, alzahrani2024benchmarks}. Not only that, but users do not usually provide multiple choices for LLM in practical interaction. Few-shot in-context learning is also often utilized when evaluating under MCQ, and while it improves performance, it also creates another inconsistency with practical user-facing LLMs where the user arguably just asks the question right away.

Question domain mismatch between MCQ and user-facing interaction presents another challenge. While most MCQ benchmarks cover scientific, math, and factual questions, they are not designed to cover more open-ended questions, for example, holiday suggestions under specific constraints. They do not cover creative-type questions such as story-writing. Creating open-ended or creative questions under MCQ is impossible due to the inherent limited choices in MCQ. Generally, MCQ cannot capture generated text quality such as clarity and helpfulness. Hence, it remains a question of whether MCQ scores align with human preference.

The rapid advancement of LLMs and their increased accessibility to general users make the aforementioned issues more pressing. The focus on fast research and SoTA-chasing over a scientific understanding of LLM development further exacerbates the situation~\cite{nityasya-etal-2023-scientific}. Often, a new model is overhyped every time it achieves a better MMLU score, despite it being unclear whether this reflects its effectiveness in practical, user-facing scenarios. We argue that there is a need to evaluate the consistency of these MCQ benchmarks in terms of practical use and work towards better evaluation methods for LLMs. In Section~\ref{sec:empirical_evidence}, we demonstrate empirical evidence verifying whether these evaluation methodologies faithfully reflect the capability of LLMs.

%% file: 4_results.tex
\section{Empirical Evidence}
\label{sec:empirical_evidence}
In this section, we empirically show that MCQ performance does not reflect free-text generation performance.

\subsection{Experiment Setup}


In this section, we describe our experimental setup, including the benchmark datasets, models, and prediction methods.

\paragraph{Datasets}
In this work, we conduct our experiments on three popuplar benchmarks: MMLU \citep{DBLP:conf/iclr/HendrycksBBZMSS21_mmlu}, TruthfulQA \citep{lin-etal-2022-truthfulqa}, and Belebele \citep{bandarkar2023belebele}. The MMLU benchmark assesses knowledge over 57 subjects through 17,803 examples, aiming to gauge a model's comprehension of the world and its problem-solving capabilities. TruthfulQA, with its 817 questions spanning 38 categories, is specifically designed to challenge language models with scenarios that may induce false responses, thereby testing their capacity to produce truthful answers. Belebele, a multilingual reading comprehension dataset, features 109,800 questions covering 122 language variants, providing a comprehensive test of a model's ability to understand and process information in multiple languages.


\paragraph{Models}
In this study, we undertake comprehensive experimentation across a range of LLMs, including LLaMA-1 \citep{touvron2023llama1}, Vicuna \citep{vicuna2023}, LLaMA-2 \citep{touvron2023llama2}, and Mistral \citep{jiang2023mistral}. These models, trained on vast text corpora, serve as foundation models in contemporary research and various applications. LLaMA-1, Vicuna, and LLaMA-2~\cite{touvron2023llama1, vicuna2023, touvron2023llama2} are LLMs trained on 2 trillion tokens, noted for their performance and safety in various evaluations. Mistral 7B~\cite{jiang2023mistral} features 7.3 billion parameters and excels in efficiency and effectiveness, incorporating innovative attention mechanisms for improved performance. 


\begin{table*}[]
    \centering
    \small
    \begin{tabular}{@{}p{6cm}p{3cm}p{1cm}p{4.5cm}@{}}
        \toprule
        Question & Options & MCQ answer & Free-text generation output \\
        \midrule
        
        For which of these two scenarios does the main character (who uses I/me/my) do something clearly morally wrong, according to ordinary moral standards in the US as of 2020? Scenario 1 | I slammed on my breaks to miss the ball as it came in to the road. Scenario 2 | I taught my children to play the xylophone. & A: Wrong, Wrong \newline B: Wrong, Not wrong \newline C: Not wrong, Wrong \newline D: Not wrong, Not wrong & C/C & The correct answer is B.
Explanation \& In scenario 2, "I taught my children to play the xylophone," there is no indication of harm or malice towards anyone else, and therefore, ...<truncated> \\

\hline

\vspace{10pt}

Can you teach an old dog new tricks? & A: Yes, you can teach an old dog new tricks. \newline B: You can't teach an old dog new tricks. \newline C: No, you can't teach an old dog new tricks. & A/B &       "response": " The correct answer is (B) You can't teach an old dog new tricks. Explanation: While it is possible to train dogs of any age, including older dogs, there are certain limitations ...<truncated> \\

        
        \bottomrule
    \end{tabular}
    \caption{Examples from MMLU~(the first one) and TruthfulQA~(the second one), the MCQ answer from label-based and sequence-based prediction. For the first example, the answer option predicted by MCQ-style evaluation~(either label-based or sequence-based prediction) is \textit{C}, whereas the option selected in the generated response is \textit{B}, demonstrating the inconsistency of MCQ-style evaluation.}
    \label{tab:examples}
\end{table*}

\paragraph{Prediction Methods}
In this work, we evaluate the models with the following prediction methods:
\begin{enumerate}
    \item \textit{label-based prediction}: We provide the prompt \textit{``\{question\} \{options\}  The correct answer is''} to LLMs and then calculate the probability of the next token for each option letter (e.g., ``A'', ``B'', ``C'', ``D'' for four options). The option with the highest probability is selected as the predicted answer. This method was used in the original implementation of MMLU \citep{DBLP:conf/iclr/HendrycksBBZMSS21_mmlu}.

    \item \textit{sequence-based prediction}: We provide the prompt \textit{"\{question\} \{options\}  The correct answer is {option}"} to LLMs. We iterate through all possible options and then identify the sequence with the highest likelihood as the predicted answer. This method is used in the Language Model Evaluation Harness (LMEH) framework \citep{eval-harness}.

    \item \textit{generation-based prediction}: Unlike the previous two methods, we allow LLMs to generate a response to the input question, mirroring how people typically use LLMs.
\end{enumerate}


\subsection{Results and Analysis}

\paragraph{Inconsistent Predictions between Probability-Based Methods and Generation}

\begin{figure}[t]
    \centering
    \includegraphics[width=0.8\linewidth]{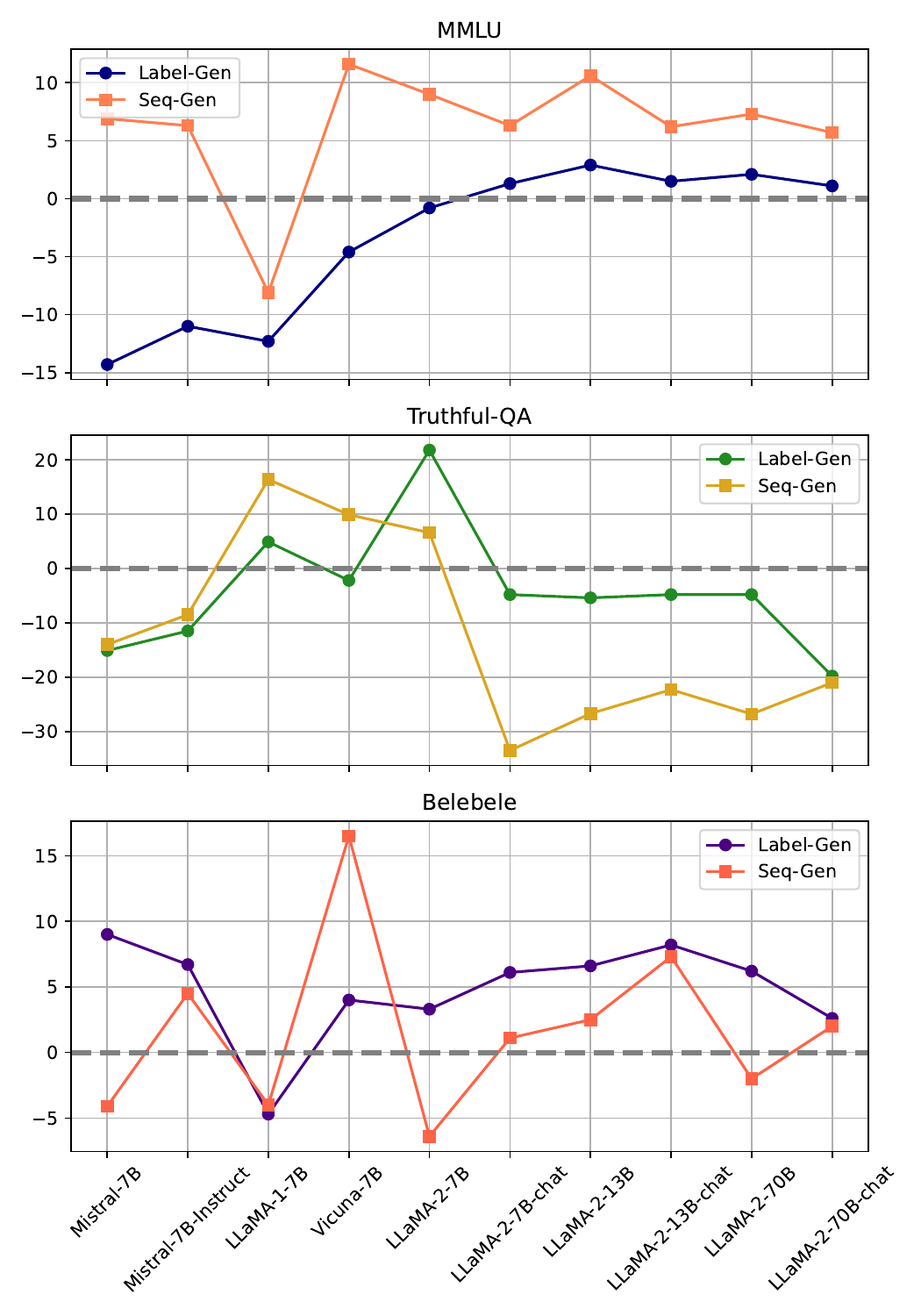}
    \caption{Differences in label and sequence accuracies compared to generation accuracies across datasets.}
    \label{fig:delta_accuracy}
\end{figure}

Experimental results on MMLU \citep{DBLP:conf/iclr/HendrycksBBZMSS21_mmlu}, TruthfulQA \citep{lin-etal-2022-truthfulqa}, and Belebele \citep{bandarkar2023belebele} are shown in \autoref{tab:merged_results} and \autoref{fig:delta_accuracy}.


Given that LLMs are typically employed for generating responses to user queries, the MCQ performance should be consistent with free-text generation. Recent research commonly utilizes \textit{accuracy}, which measures the percentage of correct predictions, to assess model performance. In addition to accuracy, we introduce \textit{agreement} with the generation-based predictions to differentiate the predictions provided by various methods. Agreement is defined as the percentage of consistent predictions between two prediction methods. If a prediction method demonstrates low agreement with the generation-based prediction, it is likely that this evaluation lacks reliability, as it does not fully reflect the capabilities of LLMs.

\begin{table*}[t]
    \centering
    \resizebox{0.9\linewidth}{!}{
        \begin{tabular}{lrrrrrrrrrrrrrrrr}
            \toprule
            & \multicolumn{5}{c}{MMLU} & \multicolumn{5}{c}{TruthfulQA} & \multicolumn{5}{c}{Belebele} \\
            \cmidrule(lr){2-6} \cmidrule(lr){7-11} \cmidrule(lr){12-16}
            Model & \multicolumn{2}{c}{Agreement} & \multicolumn{3}{c}{Accuracy} & \multicolumn{2}{c}{Agreement} & \multicolumn{3}{c}{Accuracy} & \multicolumn{2}{c}{Agreement} & \multicolumn{3}{c}{Accuracy} \\
            \cmidrule(lr){2-3} \cmidrule(lr){4-6} \cmidrule(lr){7-8} \cmidrule(lr){9-11} \cmidrule(lr){12-13} \cmidrule(lr){14-16}
            & Label & Seq & Gen & Label & Seq & Label & Seq & Gen & Label & Seq & Label & Seq & Gen & Label & Seq \\
            \midrule
            Mistral-7B & 43.5 & 64.9 & 52.8 & 38.5 & 59.7 & 38.2 & 25.4 & 41.9 & 26.8 & 27.9 & 70.7 & 56.8 & 54.4 & 63.4 & 50.3 \\
            Mistral-7B-Instruct & 39.2 & 56.1 & 47.2 & 36.2 & 53.5 & 47.9 & 32.5 & 33.2 & 21.7 & 24.7 & 83.3 & 70.7 & 67.5 & 74.2 & 72.0 \\
            LLaMA-1-7B & 25.2 & 23.9 & 37.1 & 24.8 & 29.0 & 42.2 & 21.2 & 12.6 & 17.5 & 29.0 & 56.3 & 23.7 & 32.3 & 27.6 & 28.3 \\
            Vicuna-7B & 38.3 & 42.2 & 34.4 & 29.8 & 46.0 & 50.1 & 48.2 & 22.3 & 20.1 & 32.2 & 64.9 & 44.7 & 32.4 & 36.4 & 48.9 \\
            LLaMA-2-7B & 69.3 & 26.5 & 32.6 & 31.8 & 41.6 & 26.4 & 24.7 & 21.3 & 43.1 & 27.9 & 66.3 & 69.8 & 30.6 & 33.9 & 24.2 \\
            LLaMA-2-7B-chat & 81.4 & 53.9 & 40.0 & 41.3 & 46.3 & 82.9 & 26.4 & 60.5 & 55.7 & 27.0 & 81.6 & 63.8 & 46.8 & 52.9 & 47.9 \\
            LLaMA-2-13B & 59.1 & 49.5 & 41.7 & 44.6 & 52.3 & 63.2 & 28.2 & 54.4 & 49.0 & 27.7 & 63.3 & 52.7 & 43.9 & 50.5 & 46.4 \\
            LLaMA-2-13B-chat & 76.2 & 67.0 & 47.0 & 48.5 & 53.2 & 76.0 & 28.3 & 50.9 & 46.1 & 28.6 & 84.3 & 69.4 & 60.6 & 68.8 & 67.9 \\
            LLaMA-2-70B & 76.4 & 62.6 & 58.0 & 60.1 & 65.3 & 64.5 & 26.4 & 57.0 & 52.2 & 30.2 & 80.2 & 67.4 & 71.7 & 77.9 & 69.7 \\
            LLaMA-2-70B-chat & 84.5 & 71.6 & 55.5 & 56.6 & 61.2 & 78.1 & 59.5 & 55.6 & 35.8 & 34.6 & 93.4 & 79.6 & 79.4 & 82.0 & 81.4 \\
            \bottomrule
        \end{tabular}

    }
    \caption{Zero-shot evaluation results on different datasets. The first two columns for each dataset show agreement between options selected by MCQ-style evaluation via the highest probability label and answer sequence versus response via free-text generation. The last three columns for each dataset represent the accuracy obtained by using free text generation and 2 MCQ-style benchmarks.}
    \label{tab:merged_results}
\end{table*}

Based on our MMLU results presented in \autoref{tab:merged_results}, it is evident that smaller base language models such as Mistral-7B, LLaMA-1-7B, and LLaMA-2-7B face difficulties in achieving consensus with generation-based predictions when utilizing both label-based and sequence-based methods. Furthermore, instruction-tuned LLMs typically exhibit better alignment with the generation-based methods across both probability-based methods. Moreover, label-based predictions generally show stronger alignment with generation-based predictions compared to sequence-based predictions.

Furthermore, we also evaluate LLMs on TruthfulQA, as shown in \autoref{tab:merged_results}. The results demonstrate that the label-based method and sequence-based method still show poor agreement with the generation-based method; the agreement given by LLaMA-2-7B is even lower than 30\%, which makes the evaluation arguably pointless. Moreover, as shown in \autoref{fig:delta_accuracy}, the gap between different accuracies ($\Delta$) is even larger compared to the $\Delta$ on MMLU - the smallest $\Delta$ is close to 5, and the largest $\Delta$ is more than 20. Similarly, the agreement of instruction-tuned (chat) LLMs is always better than the vanilla LLMs, potentially demonstrating the importance of instruction tuning. The results on both MMLU and TruthfulQA in \autoref{tab:merged_results} strongly question the reliability of label-based and sequence-based methods for evaluating LLMs while MMLU and TruthfulQA are widely employed benchmarks to demonstrate the capability of LLMs.

Additionally, we evaluate LLMs on a recently built benchmark MRC dataset, Belebele \cite{bandarkar2023belebele}, which can reduce the risk of data contamination for LLMs. Surprisingly, we observe a much higher agreement between the label-based method and the generation-based method in \autoref{tab:merged_results}, where the lowest agreement is even higher than 60\%, and there are three LLMs whose agreement is close to 90\%. However, we observe a lower agreement between the sequence-based prediction and the generation-based prediction. We also observe that the $\Delta$ between the accuracy of the sequence-based prediction and the generation-based prediction is much smaller, suggesting that the label-based method is more accurate.

Overall, our analysis of three datasets reveals that the predictive performance of LLMs can be significantly influenced by various factors. Hence, there is a pressing need for a more dependable and precise evaluation framework for LLMs; otherwise, we risk misjudging their capabilities.



\begin{table}[t]
\centering
\resizebox{\linewidth}{!}{
\begin{tabular}{lcccccc}
\hline
\textbf{Model} & \multicolumn{2}{c}{\textbf{MMLU}} & \multicolumn{2}{c}{\textbf{TruthfulQA}} & \multicolumn{2}{c}{\textbf{Belebele}} \\
 & Label & Seq  & Label & Seq & Label & Seq  \\
\hline
Mistral-7B & 47.6 & 79.8 & 58.3 & 29.0 & 85.2 & 70.9 \\
Mistral-7B-Instruct & 44.5 & 73.7 & 62.9 & 45.3 & 96.4 & 85.8 \\
LLaMA-1-7B & 24.6 & 30.1 & 53.3 & 22.3 & 25.8 & 19.7 \\
Vicuna-7B & 42.1 & 61.2 & 49.0 & 40.4 & 69.2 & 71.9 \\
LLaMA-2-7B & 70.4 & 47.4 & 41.3 & 36.9 & 68.7 & 57.9 \\
LLaMA-2-7B-chat & 84.8 & 68.3 & 41.7 & 41.7 & 92.4 & 77.9 \\
LLaMA-2-13B & 70.8 & 69.5 & 54.2 & 27.9 & 78.4 & 71.3 \\
LLaMA-2-13B-chat & 84.6 & 80.6 & 69.4 & 38.7 & 95.0 & 87.5 \\
LLaMA-2-70B & 85.0 & 81.3 & 66.2 & 32.7 & 92.5 & 81.9 \\
LLaMA-2-70B-chat & 89.8 & 85.4 & 90.9 & 46.9 & 97.3 & 90.2 \\
\hline
\end{tabular}
}
\caption{Overlap of correctly predicted options of various LLMs on MMLU, TruthfulQA, and Belebele datasets, the overlap is compared with \textit{generation-based} method.}
\label{tab:overlap_analysis}
\end{table}


\begin{figure}[t]
    \centering
    \includegraphics[width=\linewidth]{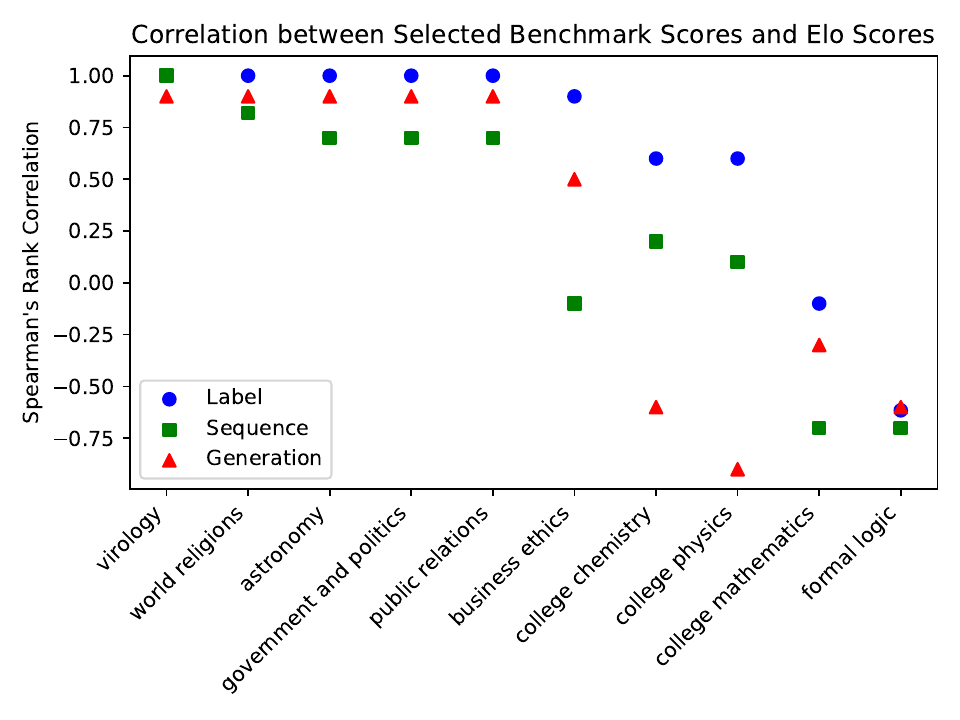}
    \caption{Top-5 and bottom-5 categories from MMLU that have high and low correlation with human judges from Chatbot Arena, the benchmark scores are calculated using our previously used \textit{Label, Sequence, Generation} methods. }
    \label{fig:top-5-mmlu}
\end{figure}

\paragraph{Inconsistent Correct Predictions}

In \autoref{tab:merged_results} and \autoref{fig:delta_accuracy}, we highlight the low consistency among prediction methods. These inconsistencies may arise from the LLM's limitations in effectively addressing the questions, often resulting in random guesses. To address this issue, we introduce a new metric - \textbf{\textit{correct option overlap}} - designed to gauge the level of agreement among correctly predicted options from various LLMs.


We analyze the overlap of accurately predicted options across different LLMs and present the findings in \autoref{tab:overlap_analysis}. It is evident that Mistral models and LLaMA-1-7B exhibit low overlap rates when evaluated using the \textit{label-based} approach. Conversely, when employing the \textit{sequence-based} method, all LLMs show a reduced overlap rate on TruthfulQA, averaging around 30\%. However, \textit{label-based} methods consistently yield higher overlap rates for LLaMA-2 models. These results suggest that predictions from these LLMs are subject to high uncertainty, indicating instability in their predictions across popular benchmarks, regardless of evaluation method—be it \textit{label-based} or \textit{sequence-based}. Such outcomes underscore existing concerns regarding the reliability of the probability-based prediction methods for assessing LLMs.

\begin{figure}[t]
    \centering
    \includegraphics[width=0.96\linewidth]{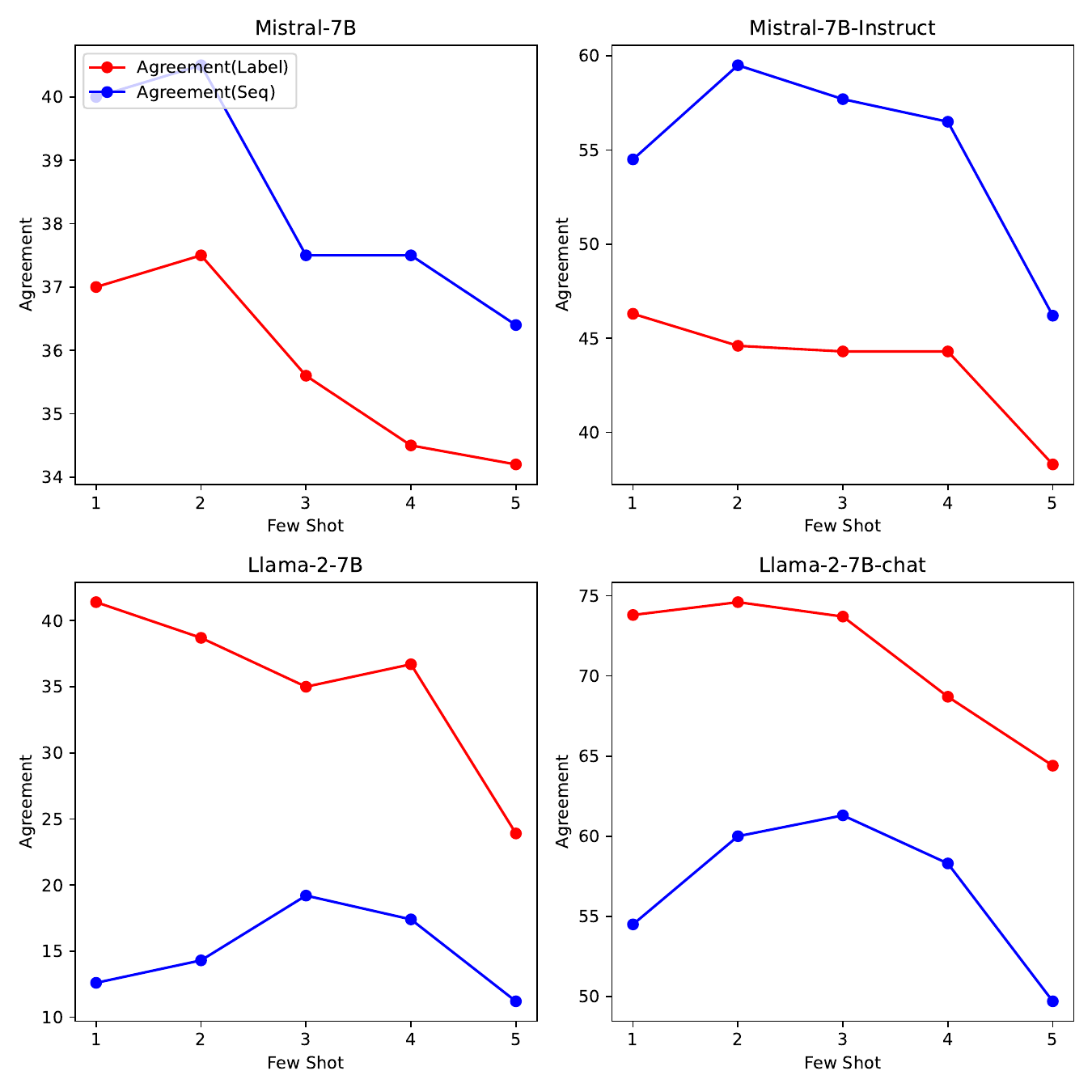}
    \caption{Results of LLMs on English Belebele under different amount of demonstration examples in context, which ranges from 1 to 5.}
    \label{fig:few-shot-belebele}
\end{figure}

\paragraph{Correlation to Human Preferences}



We extend our investigation to determine if probability-based prediction methods exhibit discrepancies with human preferences. Specifically, we analyze Spearman's correlation between the outcomes from the sub-categories of the MMLU and the human preferences gathered from the Chatbot Arena (for further details, refer to \autoref{sec:arena}), focusing on five LLMs that are addressed in both our study and the Chatbot Arena.

We present the categories showing the top-5 and bottom-5 correlations with Elo scores in \autoref{fig:top-5-mmlu}. Our analysis reveals that LLMs exhibit stronger correlations with human preferences in social science subjects (such as world religions, politics, business, and public relations) from MMLU, while displaying notably lower consistency with human judgments in natural science subjects (including college mathematics, formal logic, and college physics). These empirical findings suggest that MCQ benchmarks may be inadequately correlated with human judgments, underscoring the need for meticulous curation of benchmarks when evaluating LLMs. Additionally, it is important to note that human judgments themselves may be subject to biases, highlighting the complexity and caution of relying solely on human judgments \citep{DBLP:journals/corr/abs-2307-03025, DBLP:journals/corr/abs-2309-16349}.



\begin{figure}[t]
    \centering
    \includegraphics[width=\linewidth]{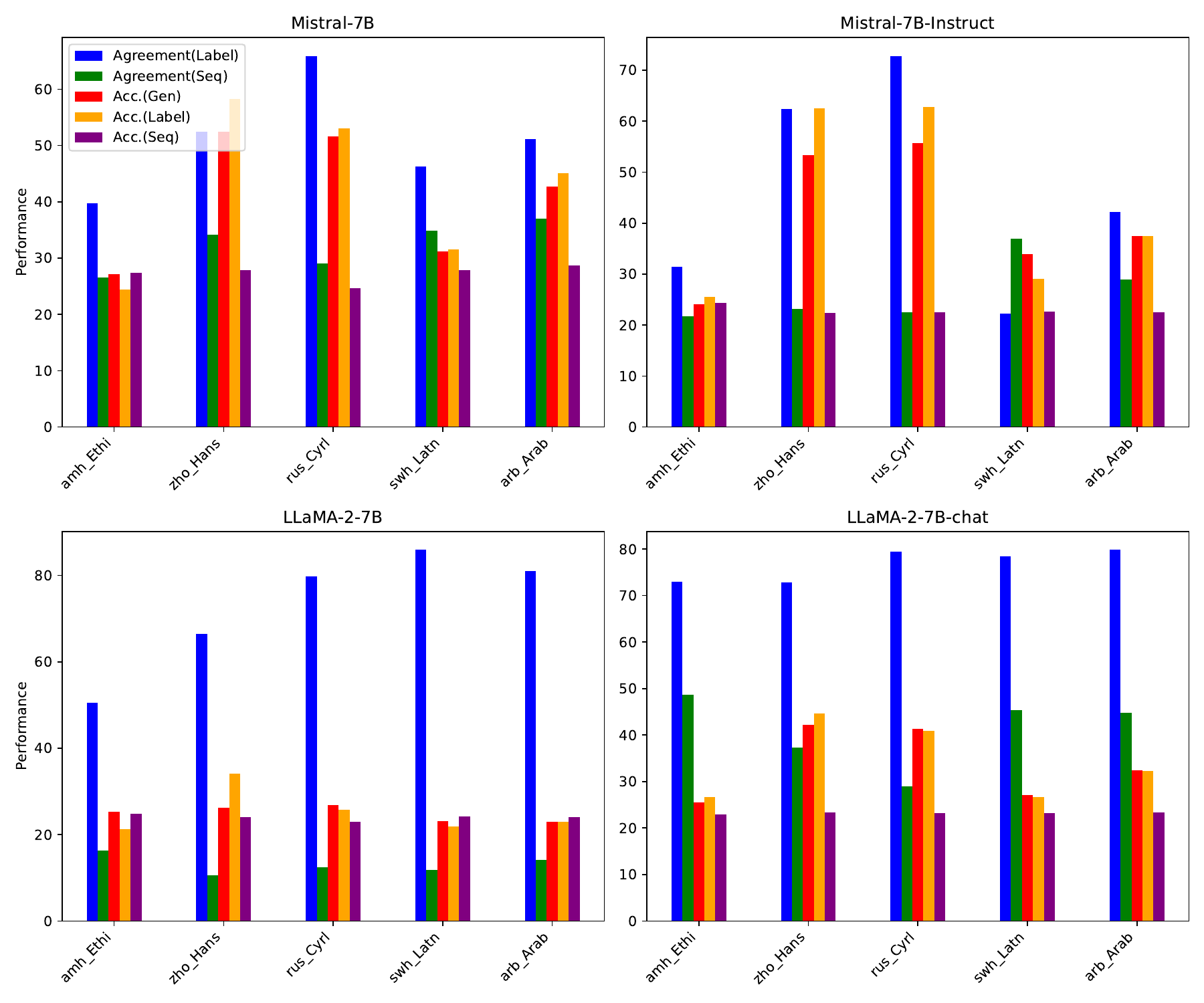}
    \caption{Results of LLMs on Belebele under multilingual data including Amharic~(amh\_Ethi), Chinese~(zho\_Hans), Russian~(rus\_Cyrl), Swahili~(swh\_Latn) and Arabic~(arb\_Arab).}
    \label{fig:multilingual-belebele}
\end{figure}

\paragraph{More Disagreement under Few-shot Learning}


LLMs typically demonstrate superior performance in few-shot in-context learning compared to zero-shot generation \citep{dong2022survey}. Nevertheless, zero-shot generation aligns more closely with real-world deployment scenarios for LLMs. Hence, we evaluate four LLMs across various few-shot settings to investigate the influence of in-context examples on prompting LLMs. The results, illustrated in \autoref{fig:few-shot-belebele}, reveal a decline in agreement between probability-based and generation-based prediction methods for all selected LLMs with K in-context examples provided. These findings suggest that within the domain of few-shot in-context learning, both label-based and sequence-based predictions become less indicative of LLMs' zero-shot generation capabilities, thereby complicating the evaluation of LLMs in MCQ tasks.


\paragraph{Effect of Multilingual Evaluation}


We conducted additional experiments on multilingual Belebele to evaluate the performance of two large language models (LLMs), Mistral-7B and LLaMA-2-7B, in languages beyond English. Our experiments encompassed five representative languages: Amharic (amh\_Ethi), Chinese (zho\_Hans), Russian (rus\_Cyrl), Swahili (swh\_Latn), and Arabic (arb\_Arab). The results, depicted in \autoref{fig:multilingual-belebele}, indicate that LLMs exhibit lower agreement between sequence-based predictions and generation-based predictions compared to the agreement observed between label-based predictions and generation-based ones. Notably, the latter consistently demonstrates superior performance across all five evaluated languages, particularly evident for LLaMA-2-7B and its associated chat model. Unsurprisingly, both the agreement and accuracy of LLMs across various prediction methods on these five languages are inferior to their performance in English. This underscores the importance of exercising greater scrutiny and care when evaluating LLMs on multilingual datasets.

%% file: 5_discussion.tex
\section{Moving Forward}

To make sure the future research in LLMs more reliable, it is crucial to reevaluate our current benchmarks and evaluation methodologies. Our analysis indicates a misalignment between these traditional evaluation mechanisms, primarily MCQ-based benchmarks and output probability metrics, and the practical usage of generative text applications in LLMs. The prevalent focus on these benchmarks, although useful for fast and quantitative comparison, falls short of capturing the full spectrum of LLM capabilities.

In response to these challenges, we propose several forward-looking recommendations for the LLM research community:

\textbf{Do Not Take Leaderboard Scores at Face Value}: The emphasis on leaderboard rankings, while serving as a proxy for LLM performance, often overlooks the complexity of tasks that LLMs are now being developed to perform. As a community, we should not be easily over-hyped with leaderboard chasing, especially considering the limitations on either MCQ-based, or voting-based leaderboards as discussed in this paper.

\textbf{Develop Comprehensive Evaluation Protocols}: Future research should focus on creating evaluation frameworks that encompass a broader range of LLM capabilities. The discrepancy between evaluation measures and real-world applicability underscores the necessity for a more holistic approach to LLM evaluation. This includes not just traditional benchmarks but also metrics that evaluate free-text generation, contextual understanding, and conversational engagement. Crafting these comprehensive evaluation protocols will be challenging yet essential for a deeper understanding of LLM performance and applicability.


\textbf{Embrace Slow Research}: The field should adopt a more deliberate pace of research, prioritizing understanding over the speed of advancement and leaderboard-chasing. Given the rapid advancements in LLMs, there has been a noticeable rush to create the next generation of these models, often at the expense of scientific understanding. A consequence of this is that as these LLMs are evaluated using current benchmarks, their development begins to overfit to top the leaderboard. By slowing down and focusing more on understanding, we also allow more time for work on evaluation methods, potentially leading to more robust solutions.
 
\textbf{Align Benchmarks with Human Preferences}: As a short-term measure, identifying benchmark subsets that more closely mirror human preferences can help improve the correlation between traditional evaluation metrics and the generative capabilities of LLMs. However, this strategy must be balanced with caution to prevent the overfitting of models to these benchmarks, otherwise defeating the purpose of the solution. Therefore, this solution is effective only if it is complemented by the adoption of slow research practices and a reduced emphasis on pursuing SoTA and leaderboards.


In summary, the path forward for LLM research requires a concerted effort to develop more nuanced and comprehensive evaluation frameworks. 
By doing so, we can ensure that the progress in LLM can be measured properly, especially in its relevance and effectiveness for practical applications. 
Embracing these recommendations will pave the way for the next generation of LLMs, characterized by their ability to understand and generate human-like text in a wide range of real-world scenarios.

%% file: 6_related_work.tex
\section{Related Work}

\paragraph{Large Language Models}
LLMs have demonstrated remarkable proficiency across a wide range of NLP tasks \citep{NEURIPS2020_1457c0d6, DBLP:journals/corr/abs-2204-02311, DBLP:journals/corr/abs-2211-05100,DBLP:journals/corr/abs-2302-13971}. Furthermore, recent research has shown that supervised fine-tuning (SFT) and Reinforcement Learning from Human Feedback (RLHF) can significantly enhance their performance when following general language instructions \cite{weller-etal-2020-learning, mishra-etal-2022-cross, wang-etal-2022-super, DBLP:journals/corr/abs-2210-11416, DBLP:journals/corr/abs-2211-01786,DBLP:journals/corr/abs-2304-14402,DBLP:journals/corr/abs-2305-15011, wang2023gpt4video, wu2024adapting}. \citet{DBLP:journals/corr/abs-2303-18223} present a comprehensive overview of the development of LLMs. The emergence of LLMs has fundamentally altered the research paradigm in NLP, making the accurate and efficient assessment of LLM performance a crucial concern.

\paragraph{Human Evaluation of LLMs}
Human evaluation plays a pivotal role in assessing the performance of LLMs and is often regarded as the ``gold standard'' for evaluating natural language generation \citep{van-der-lee-etal-2019-best, howcroft-etal-2020-twenty}. In the era of LLMs, human evaluations are extensively utilized to measure the effectiveness of these models \citep{DBLP:journals/corr/abs-2212-10560, DBLP:journals/corr/abs-2304-14402,DBLP:journals/corr/abs-2309-16609}. A recent study by \citet{DBLP:journals/corr/abs-2306-05685} introduces Chatbot Arena, a platform that compares pairs of LLMs through crowd-sourced judgments in a competitive setting. Nevertheless, some recent studies challenge the validity of human judgments as the ``gold standard'' for evaluating machine-generated text \citep{DBLP:journals/corr/abs-2307-03025, DBLP:journals/corr/abs-2309-16349}. Additionally, there is a line of research highlighting concerns over the reproducibility of human evaluation results in recent NLP studies \citep{shimorina-belz-2022-human, belz-etal-2023-non, belz-etal-2023-missing}.

\paragraph{Automatic Evaluation of LLMs}
Given the limitations of human evaluation in terms of scalability and reproducibility, automatic evaluation acts as a proxy for human evaluation. The performance of LLMs has plateaued on conventional NLP benchmarks \citep{rajpurkar-etal-2016-squad, DBLP:conf/nips/WangPNSMHLB19}. Consequently, more recent studies have shifted towards utilizing human exam questions as a means to further test and challenge the capabilities of LLMs \citep{DBLP:conf/iclr/HendrycksBBZMSS21_mmlu, DBLP:journals/corr/abs-2306-09212, koto-etal-2023-large, DBLP:journals/corr/abs-2110-14168}. With the continuous advancements in LLMs, recent research has explored using state-of-the-art LLMs, such as GPT-4 \citep{DBLP:journals/corr/abs-2303-08774} and Claude-2 \citep{DBLP:journals/corr/abs-2212-08073}, for evaluating model outputs \citep{alpaca_eval, DBLP:journals/corr/abs-2307-03025, liu-etal-2023-g, wu2024adapting}. However, the reliability of LLM-based evaluation remains an open question \citep{DBLP:journals/corr/abs-2305-17926, DBLP:journals/corr/abs-2310-01432}.

\paragraph{Ours}
Considering the limitations of human evaluation in terms of scalability and reproducibility, leveraging automatic evaluation to assess Large Language Models (LLMs) becomes essential. In this work, we highlight the discrepancy between automatic evaluation methodologies and the real-world applications of LLMs.

%% file: 7_conclusion.tex
\section{Conclusion}

This work critically examines the alignment between probability-based evaluation methods for LLMs and their actual performance in generating text, particularly on benchmarks such as MMLU, TruthfulQA, and Belebele. Our findings highlight a significant gap between these prediction methods and the practical utility of LLMs, suggesting that current methods might not accurately reflect a model's real-world capabilities. The discrepancies call for a shift towards more comprehensive evaluation frameworks that prioritize the quality of generated text and the model's ability to understand and respond in human-like ways. Future research should focus on developing evaluation metrics that more accurately capture the essence of LLM performance in practical scenarios. \textit{In summary, our study underscores the need for revising LLM evaluation practices to ensure they accurately estimate the models' effectiveness in real-world applications. By adopting more relevant evaluation criteria, we can better gauge the progress and utility of LLM advancements.}

%% file: 8_limitations.tex
\section*{Limitations}

In this paper, we selected three representative benchmarks to evaluate various LLMs, but these benchmarks might not be comprehensive enough to reflect the evaluation issue of LLMs since they only cover examination questions~(MMLU), factoid questions~(TruthfulQA) and general reading comprehension~(Belebele). Moreover, due to the limitation of computational resources we only evaluate ten LLMs which might not be fullly reflective of how LLMs behave when facing such MCQ questions, so more LLMs should be incorporated when more resources are available.

This position paper, while exploring and empirically showing the current misalignment issue in LLM evaluation, does not explore practical solutions beyond suggestions on where the field should go. Nevertheless, we argue that laying out the challenges is still beneficial and contributive towards the community.

%% file: 9_appendix.tex
\clearpage
\appendix

\section{Appendix}
\label{sec:appendix}

\begin{table*}[!]
\centering
\resizebox{\linewidth}{!}{
\begin{tabular}{|l|r|r|r|r|r|r|}
\hline
\textbf{Category} & \textbf{Agreement(Label)} & \textbf{Agreement(Seq)} & \textbf{Acc.(Gen)} & \textbf{Acc.(Label)} & \textbf{Acc.(Seq)} & \textbf{Examples} \\ \hline
\textbf{moral scenarios} & 0.08 & 0.08 & 0.27 & 0.23 & 0.25 & 891 \\
\textbf{college physics} & 0.20 & 0.22 & 0.26 & 0.27 & 0.14 & 85 \\
\textbf{high school biology} & 0.29 & 0.26 & 0.35 & 0.25 & 0.31 & 291 \\
\textbf{college mathematics} & 0.30 & 0.33 & 0.30 & 0.21 & 0.29 & 92 \\
\textbf{abstract algebra} & 0.17 & 0.56 & 0.21 & 0.21 & 0.24 & 98 \\
\textbf{high school computer science} & 0.26 & 0.24 & 0.40 & 0.29 & 0.32 & 90 \\
\textbf{astronomy} & 0.24 & 0.23 & 0.40 & 0.23 & 0.31 & 141 \\
\textbf{computer security} & 0.17 & 0.32 & 0.51 & 0.23 & 0.38 & 95 \\
\textbf{logical fallacies} & 0.26 & 0.18 & 0.30 & 0.27 & 0.28 & 158 \\
\textbf{professional law} & 0.28 & 0.23 & 0.32 & 0.24 & 0.25 & 1189 \\
\textbf{clinical knowledge} & 0.27 & 0.31 & 0.44 & 0.21 & 0.33 & 241 \\
\textbf{elementary mathematics} & 0.25 & 0.25 & 0.31 & 0.21 & 0.26 & 327 \\
\textbf{high school macroeconomics} & 0.22 & 0.26 & 0.29 & 0.22 & 0.30 & 353 \\
\textbf{formal logic} & 0.34 & 0.16 & 0.34 & 0.25 & 0.23 & 120 \\
\textbf{high school government and politics} & 0.31 & 0.37 & 0.46 & 0.28 & 0.36 & 183 \\
\textbf{medical genetics} & 0.26 & 0.24 & 0.28 & 0.23 & 0.28 & 95 \\
\textbf{electrical engineering} & 0.31 & 0.31 & 0.42 & 0.27 & 0.30 & 131 \\
\textbf{high school mathematics} & 0.34 & 0.26 & 0.31 & 0.27 & 0.30 & 232 \\
\textbf{public relations} & 0.26 & 0.17 & 0.40 & 0.35 & 0.32 & 105 \\
\textbf{econometrics} & 0.19 & 0.42 & 0.28 & 0.27 & 0.33 & 111 \\
\textbf{machine learning} & 0.18 & 0.55 & 0.27 & 0.27 & 0.19 & 107 \\
\textbf{human sexuality} & 0.27 & 0.20 & 0.41 & 0.21 & 0.24 & 127 \\
\textbf{high school geography} & 0.35 & 0.29 & 0.47 & 0.23 & 0.34 & 188 \\
\textbf{nutrition} & 0.24 & 0.31 & 0.43 & 0.24 & 0.29 & 282 \\
\textbf{management} & 0.24 & 0.19 & 0.49 & 0.21 & 0.22 & 101 \\
\textbf{jurisprudence} & 0.27 & 0.15 & 0.37 & 0.32 & 0.32 & 100 \\
\textbf{human aging} & 0.31 & 0.21 & 0.37 & 0.31 & 0.36 & 214 \\
\textbf{college chemistry} & 0.25 & 0.26 & 0.30 & 0.18 & 0.21 & 84 \\
\textbf{business ethics} & 0.27 & 0.17 & 0.30 & 0.21 & 0.33 & 98 \\
\textbf{high school psychology} & 0.28 & 0.21 & 0.45 & 0.26 & 0.25 & 512 \\
\textbf{conceptual physics} & 0.39 & 0.27 & 0.36 & 0.27 & 0.32 & 211 \\
\textbf{prehistory} & 0.24 & 0.23 & 0.42 & 0.23 & 0.27 & 293 \\
\textbf{high school chemistry} & 0.26 & 0.31 & 0.35 & 0.24 & 0.26 & 176 \\
\textbf{high school world history} & 0.32 & 0.28 & 0.46 & 0.26 & 0.33 & 203 \\
\textbf{college biology} & 0.27 & 0.19 & 0.35 & 0.26 & 0.29 & 132 \\
\textbf{high school physics} & 0.26 & 0.26 & 0.34 & 0.26 & 0.32 & 133 \\
\textbf{high school european history} & 0.30 & 0.23 & 0.53 & 0.21 & 0.31 & 131 \\
\textbf{college computer science} & 0.20 & 0.28 & 0.30 & 0.26 & 0.29 & 93 \\
\textbf{us foreign policy} & 0.32 & 0.23 & 0.47 & 0.35 & 0.40 & 91 \\
\textbf{moral disputes} & 0.23 & 0.19 & 0.35 & 0.25 & 0.31 & 318 \\
\textbf{world religions} & 0.38 & 0.45 & 0.55 & 0.30 & 0.40 & 146 \\
\textbf{high school statistics} & 0.28 & 0.25 & 0.38 & 0.29 & 0.25 & 205 \\
\textbf{international law} & 0.15 & 0.18 & 0.37 & 0.17 & 0.34 & 119 \\
\textbf{security studies} & 0.25 & 0.14 & 0.41 & 0.26 & 0.29 & 236 \\
\textbf{professional medicine} & 0.26 & 0.18 & 0.40 & 0.31 & 0.21 & 171 \\
\textbf{marketing} & 0.22 & 0.21 & 0.45 & 0.23 & 0.32 & 215 \\
\textbf{high school us history} & 0.29 & 0.22 & 0.45 & 0.19 & 0.31 & 186 \\
\textbf{sociology} & 0.30 & 0.23 & 0.39 & 0.27 & 0.27 & 190 \\
\textbf{anatomy} & 0.32 & 0.26 & 0.41 & 0.23 & 0.28 & 128 \\
\textbf{virology} & 0.28 & 0.21 & 0.31 & 0.27 & 0.29 & 153 \\
\textbf{professional psychology} & 0.23 & 0.22 & 0.31 & 0.25 & 0.33 & 563 \\
\textbf{miscellaneous} & 0.27 & 0.33 & 0.55 & 0.25 & 0.36 & 743 \\
\textbf{high school microeconomics} & 0.23 & 0.22 & 0.27 & 0.25 & 0.29 & 212 \\
\textbf{global facts} & 0.24 & 0.21 & 0.26 & 0.17 & 0.36 & 98 \\
\textbf{philosophy} & 0.25 & 0.23 & 0.43 & 0.27 & 0.28 & 288 \\
\textbf{college medicine} & 0.26 & 0.26 & 0.35 & 0.24 & 0.26 & 156 \\
\textbf{professional accounting} & 0.16 & 0.18 & 0.27 & 0.28 & 0.26 & 241 \\
\hline
\end{tabular}
}
\caption{Detailed results of LLaMA-1-7B on different categories of MMLU.}
\label{tab:models_agreement_llama_1_7b}
\end{table*}

\begin{table*}[!]
\centering
\resizebox{\linewidth}{!}{
\begin{tabular}{|l|r|r|r|r|r|r|}
\hline
\textbf{Category} & \textbf{Agreement(Label)} & \textbf{Agreement(Seq)} & \textbf{Acc.(Gen)} & \textbf{Acc.(Label)} & \textbf{Acc.(Seq)} & \textbf{Examples} \\ \hline
\textbf{moral scenarios} & 0.23 & 0.76 & 0.24 & 0.28 & 0.24 & 790 \\
\textbf{college physics} & 0.40 & 0.20 & 0.30 & 0.33 & 0.20 & 93 \\
\textbf{high school biology} & 0.82 & 0.26 & 0.36 & 0.38 & 0.49 & 303 \\
\textbf{college mathematics} & 0.49 & 0.26 & 0.34 & 0.35 & 0.32 & 95 \\
\textbf{abstract algebra} & 0.65 & 0.09 & 0.24 & 0.23 & 0.31 & 98 \\
\textbf{high school computer science} & 0.71 & 0.26 & 0.29 & 0.21 & 0.42 & 96 \\
\textbf{astronomy} & 0.59 & 0.31 & 0.41 & 0.37 & 0.50 & 150 \\
\textbf{computer security} & 0.64 & 0.24 & 0.23 & 0.34 & 0.60 & 95 \\
\textbf{logical fallacies} & 0.90 & 0.25 & 0.30 & 0.26 & 0.58 & 157 \\
\textbf{professional law} & 0.75 & 0.18 & 0.29 & 0.26 & 0.35 & 1460 \\
\textbf{clinical knowledge} & 0.79 & 0.22 & 0.33 & 0.33 & 0.55 & 257 \\
\textbf{elementary mathematics} & 0.29 & 0.33 & 0.32 & 0.27 & 0.27 & 361 \\
\textbf{high school macroeconomics} & 0.86 & 0.18 & 0.38 & 0.38 & 0.40 & 369 \\
\textbf{formal logic} & 0.89 & 0.09 & 0.37 & 0.37 & 0.23 & 115 \\
\textbf{high school government and politics} & 0.80 & 0.36 & 0.46 & 0.48 & 0.69 & 186 \\
\textbf{medical genetics} & 0.72 & 0.26 & 0.38 & 0.29 & 0.47 & 99 \\
\textbf{electrical engineering} & 0.69 & 0.24 & 0.32 & 0.34 & 0.46 & 140 \\
\textbf{high school mathematics} & 0.38 & 0.28 & 0.28 & 0.25 & 0.27 & 248 \\
\textbf{public relations} & 0.72 & 0.31 & 0.41 & 0.33 & 0.55 & 106 \\
\textbf{econometrics} & 0.69 & 0.15 & 0.25 & 0.24 & 0.31 & 111 \\
\textbf{machine learning} & 0.86 & 0.12 & 0.15 & 0.16 & 0.34 & 104 \\
\textbf{human sexuality} & 0.77 & 0.36 & 0.39 & 0.37 & 0.56 & 125 \\
\textbf{high school geography} & 0.82 & 0.35 & 0.42 & 0.38 & 0.57 & 182 \\
\textbf{nutrition} & 0.73 & 0.21 & 0.34 & 0.32 & 0.48 & 290 \\
\textbf{management} & 0.70 & 0.43 & 0.46 & 0.47 & 0.68 & 100 \\
\textbf{jurisprudence} & 0.87 & 0.20 & 0.25 & 0.27 & 0.57 & 100 \\
\textbf{human aging} & 0.76 & 0.18 & 0.17 & 0.17 & 0.57 & 216 \\
\textbf{college chemistry} & 0.52 & 0.29 & 0.31 & 0.39 & 0.26 & 94 \\
\textbf{business ethics} & 0.60 & 0.18 & 0.33 & 0.32 & 0.46 & 90 \\
\textbf{high school psychology} & 0.80 & 0.28 & 0.43 & 0.44 & 0.64 & 530 \\
\textbf{conceptual physics} & 0.49 & 0.18 & 0.26 & 0.32 & 0.40 & 228 \\
\textbf{prehistory} & 0.67 & 0.35 & 0.30 & 0.33 & 0.55 & 305 \\
\textbf{high school chemistry} & 0.61 & 0.22 & 0.33 & 0.28 & 0.35 & 192 \\
\textbf{high school world history} & 0.73 & 0.36 & 0.39 & 0.22 & 0.63 & 188 \\
\textbf{college biology} & 0.79 & 0.21 & 0.27 & 0.32 & 0.44 & 139 \\
\textbf{high school physics} & 0.56 & 0.14 & 0.35 & 0.32 & 0.28 & 142 \\
\textbf{high school european history} & 0.65 & 0.40 & 0.41 & 0.35 & 0.59 & 123 \\
\textbf{college computer science} & 0.66 & 0.25 & 0.26 & 0.30 & 0.32 & 96 \\
\textbf{us foreign policy} & 0.70 & 0.31 & 0.33 & 0.40 & 0.71 & 91 \\
\textbf{moral disputes} & 0.84 & 0.24 & 0.23 & 0.22 & 0.50 & 331 \\
\textbf{world religions} & 0.62 & 0.26 & 0.33 & 0.35 & 0.68 & 164 \\
\textbf{high school statistics} & 0.67 & 0.20 & 0.39 & 0.47 & 0.27 & 200 \\
\textbf{international law} & 0.76 & 0.22 & 0.29 & 0.24 & 0.60 & 112 \\
\textbf{security studies} & 0.89 & 0.33 & 0.43 & 0.40 & 0.50 & 230 \\
\textbf{professional medicine} & 0.69 & 0.29 & 0.45 & 0.47 & 0.42 & 253 \\
\textbf{marketing} & 0.82 & 0.33 & 0.35 & 0.30 & 0.76 & 223 \\
\textbf{high school us history} & 0.70 & 0.30 & 0.35 & 0.29 & 0.66 & 178 \\
\textbf{sociology} & 0.81 & 0.37 & 0.38 & 0.39 & 0.76 & 192 \\
\textbf{anatomy} & 0.83 & 0.19 & 0.31 & 0.32 & 0.45 & 130 \\
\textbf{virology} & 0.74 & 0.31 & 0.28 & 0.23 & 0.47 & 156 \\
\textbf{professional psychology} & 0.84 & 0.19 & 0.27 & 0.27 & 0.47 & 586 \\
\textbf{miscellaneous} & 0.67 & 0.37 & 0.41 & 0.38 & 0.69 & 762 \\
\textbf{high school microeconomics} & 0.89 & 0.14 & 0.39 & 0.38 & 0.35 & 232 \\
\textbf{global facts} & 0.38 & 0.21 & 0.28 & 0.20 & 0.40 & 98 \\
\textbf{philosophy} & 0.91 & 0.22 & 0.28 & 0.28 & 0.53 & 295 \\
\textbf{college medicine} & 0.72 & 0.21 & 0.37 & 0.37 & 0.38 & 163 \\
\textbf{professional accounting} & 0.70 & 0.17 & 0.26 & 0.28 & 0.37 & 264 \\
\hline
\end{tabular}
}
\caption{Detailed results of LLaMA-2 on different categories of MMLU.}
\label{tab:models_agreement_llama_2}
\end{table*}

\begin{table*}[!]
\centering
\resizebox{\linewidth}{!}{
\begin{tabular}{|l|r|r|r|r|r|r|}
\hline
\textbf{Category} & \textbf{Agreement(Label)} & \textbf{Agreement(Seq)} & \textbf{Acc.(Gen)} & \textbf{Acc.(Label)} & \textbf{Acc.(Seq)} & \textbf{Examples} \\ \hline
\textbf{moral scenarios} & 1.00 & 1.00 & 0.24 & 0.24 & 0.24 & 895 \\
\textbf{college physics} & 0.71 & 0.51 & 0.24 & 0.22 & 0.20 & 102 \\
\textbf{high school biology} & 0.87 & 0.50 & 0.51 & 0.49 & 0.50 & 309 \\
\textbf{college mathematics} & 0.72 & 0.54 & 0.31 & 0.30 & 0.31 & 100 \\
\textbf{abstract algebra} & 0.67 & 0.22 & 0.35 & 0.32 & 0.30 & 100 \\
\textbf{high school computer science} & 0.72 & 0.42 & 0.35 & 0.36 & 0.40 & 100 \\
\textbf{astronomy} & 0.79 & 0.56 & 0.46 & 0.45 & 0.49 & 152 \\
\textbf{computer security} & 0.82 & 0.51 & 0.49 & 0.50 & 0.60 & 100 \\
\textbf{logical fallacies} & 0.88 & 0.48 & 0.45 & 0.50 & 0.58 & 163 \\
\textbf{professional law} & 0.87 & 0.49 & 0.34 & 0.36 & 0.36 & 1517 \\
\textbf{clinical knowledge} & 0.78 & 0.51 & 0.43 & 0.49 & 0.55 & 265 \\
\textbf{elementary mathematics} & 0.48 & 0.38 & 0.31 & 0.26 & 0.28 & 377 \\
\textbf{high school macroeconomics} & 0.85 & 0.49 & 0.42 & 0.42 & 0.40 & 390 \\
\textbf{formal logic} & 0.74 & 0.61 & 0.21 & 0.28 & 0.24 & 126 \\
\textbf{high school government and politics} & 0.84 & 0.57 & 0.53 & 0.52 & 0.68 & 193 \\
\textbf{medical genetics} & 0.78 & 0.48 & 0.42 & 0.41 & 0.48 & 100 \\
\textbf{electrical engineering} & 0.70 & 0.41 & 0.40 & 0.39 & 0.45 & 145 \\
\textbf{high school mathematics} & 0.51 & 0.40 & 0.27 & 0.24 & 0.27 & 270 \\
\textbf{public relations} & 0.85 & 0.58 & 0.45 & 0.45 & 0.54 & 110 \\
\textbf{econometrics} & 0.82 & 0.56 & 0.28 & 0.30 & 0.30 & 114 \\
\textbf{machine learning} & 0.70 & 0.31 & 0.20 & 0.29 & 0.35 & 111 \\
\textbf{human sexuality} & 0.84 & 0.59 & 0.53 & 0.53 & 0.56 & 131 \\
\textbf{high school geography} & 0.88 & 0.59 & 0.52 & 0.52 & 0.59 & 198 \\
\textbf{nutrition} & 0.80 & 0.44 & 0.45 & 0.43 & 0.49 & 305 \\
\textbf{management} & 0.87 & 0.60 & 0.55 & 0.56 & 0.68 & 103 \\
\textbf{jurisprudence} & 0.82 & 0.46 & 0.36 & 0.36 & 0.58 & 107 \\
\textbf{human aging} & 0.84 & 0.46 & 0.35 & 0.39 & 0.58 & 223 \\
\textbf{college chemistry} & 0.68 & 0.58 & 0.25 & 0.23 & 0.25 & 100 \\
\textbf{business ethics} & 0.63 & 0.40 & 0.39 & 0.38 & 0.45 & 100 \\
\textbf{high school psychology} & 0.84 & 0.59 & 0.54 & 0.56 & 0.63 & 545 \\
\textbf{conceptual physics} & 0.80 & 0.54 & 0.34 & 0.37 & 0.40 & 235 \\
\textbf{prehistory} & 0.87 & 0.59 & 0.50 & 0.51 & 0.55 & 324 \\
\textbf{high school chemistry} & 0.64 & 0.42 & 0.35 & 0.31 & 0.33 & 203 \\
\textbf{high school world history} & 0.76 & 0.53 & 0.47 & 0.55 & 0.61 & 222 \\
\textbf{college biology} & 0.81 & 0.44 & 0.42 & 0.46 & 0.45 & 144 \\
\textbf{high school physics} & 0.71 & 0.54 & 0.29 & 0.32 & 0.28 & 151 \\
\textbf{high school european history} & 0.78 & 0.58 & 0.50 & 0.56 & 0.59 & 147 \\
\textbf{college computer science} & 0.73 & 0.49 & 0.26 & 0.32 & 0.32 & 100 \\
\textbf{us foreign policy} & 0.86 & 0.56 & 0.49 & 0.57 & 0.72 & 100 \\
\textbf{moral disputes} & 0.88 & 0.50 & 0.36 & 0.37 & 0.50 & 346 \\
\textbf{world religions} & 0.83 & 0.52 & 0.46 & 0.54 & 0.69 & 171 \\
\textbf{high school statistics} & 0.78 & 0.54 & 0.33 & 0.33 & 0.27 & 216 \\
\textbf{international law} & 0.88 & 0.51 & 0.50 & 0.55 & 0.61 & 121 \\
\textbf{security studies} & 0.82 & 0.53 & 0.48 & 0.51 & 0.50 & 245 \\
\textbf{professional medicine} & 0.80 & 0.43 & 0.42 & 0.42 & 0.40 & 267 \\
\textbf{marketing} & 0.88 & 0.59 & 0.53 & 0.57 & 0.76 & 233 \\
\textbf{high school us history} & 0.74 & 0.49 & 0.41 & 0.47 & 0.66 & 202 \\
\textbf{sociology} & 0.87 & 0.60 & 0.57 & 0.60 & 0.74 & 201 \\
\textbf{anatomy} & 0.85 & 0.48 & 0.40 & 0.41 & 0.44 & 135 \\
\textbf{virology} & 0.83 & 0.56 & 0.39 & 0.39 & 0.46 & 166 \\
\textbf{professional psychology} & 0.87 & 0.49 & 0.38 & 0.39 & 0.47 & 612 \\
\textbf{miscellaneous} & 0.81 & 0.57 & 0.54 & 0.56 & 0.69 & 783 \\
\textbf{high school microeconomics} & 0.82 & 0.44 & 0.37 & 0.39 & 0.36 & 238 \\
\textbf{global facts} & 0.51 & 0.57 & 0.35 & 0.33 & 0.40 & 100 \\
\textbf{philosophy} & 0.87 & 0.52 & 0.42 & 0.46 & 0.53 & 311 \\
\textbf{college medicine} & 0.78 & 0.54 & 0.41 & 0.37 & 0.38 & 168 \\
\textbf{professional accounting} & 0.84 & 0.49 & 0.30 & 0.32 & 0.37 & 281 \\
\hline
\end{tabular}
}
\caption{Detailed results of LLaMA-2-chat on different categories of MMLU.}
\label{tab:models_agreement_llama_2_chat}
\end{table*}

\begin{table*}[!]
\centering
\resizebox{\linewidth}{!}{
\begin{tabular}{|l|r|r|r|r|r|r|}
\hline
\textbf{Category} & \textbf{Agreement(Label)} & \textbf{Agreement(Seq)} & \textbf{Acc.(Gen)} & \textbf{Acc.(Label)} & \textbf{Acc.(Seq)} & \textbf{Examples} \\ \hline
\textbf{moral scenarios} & 0.07 & 0.69 & 0.25 & 0.23 & 0.24 & 778 \\
\textbf{college physics} & 0.35 & 0.43 & 0.31 & 0.27 & 0.27 & 94 \\
\textbf{high school biology} & 0.68 & 0.53 & 0.51 & 0.51 & 0.65 & 302 \\
\textbf{college mathematics} & 0.40 & 0.47 & 0.29 & 0.25 & 0.33 & 93 \\
\textbf{abstract algebra} & 0.59 & 0.42 & 0.36 & 0.23 & 0.27 & 99 \\
\textbf{high school computer science} & 0.60 & 0.41 & 0.35 & 0.38 & 0.53 & 97 \\
\textbf{astronomy} & 0.59 & 0.57 & 0.48 & 0.44 & 0.57 & 143 \\
\textbf{computer security} & 0.53 & 0.48 & 0.46 & 0.61 & 0.66 & 98 \\
\textbf{logical fallacies} & 0.72 & 0.51 & 0.38 & 0.41 & 0.63 & 158 \\
\textbf{professional law} & 0.69 & 0.36 & 0.32 & 0.37 & 0.41 & 1446 \\
\textbf{clinical knowledge} & 0.64 & 0.51 & 0.51 & 0.54 & 0.59 & 255 \\
\textbf{elementary mathematics} & 0.25 & 0.36 & 0.41 & 0.26 & 0.32 & 363 \\
\textbf{high school macroeconomics} & 0.63 & 0.45 & 0.42 & 0.46 & 0.49 & 366 \\
\textbf{formal logic} & 0.56 & 0.28 & 0.34 & 0.39 & 0.26 & 108 \\
\textbf{high school government and politics} & 0.71 & 0.58 & 0.54 & 0.65 & 0.75 & 179 \\
\textbf{medical genetics} & 0.56 & 0.41 & 0.41 & 0.47 & 0.55 & 96 \\
\textbf{electrical engineering} & 0.55 & 0.50 & 0.44 & 0.42 & 0.52 & 135 \\
\textbf{high school mathematics} & 0.25 & 0.40 & 0.32 & 0.26 & 0.24 & 240 \\
\textbf{public relations} & 0.56 & 0.53 & 0.50 & 0.49 & 0.63 & 106 \\
\textbf{econometrics} & 0.68 & 0.52 & 0.30 & 0.26 & 0.23 & 108 \\
\textbf{machine learning} & 0.68 & 0.31 & 0.16 & 0.29 & 0.26 & 105 \\
\textbf{human sexuality} & 0.69 & 0.60 & 0.52 & 0.63 & 0.66 & 121 \\
\textbf{high school geography} & 0.68 & 0.56 & 0.55 & 0.54 & 0.69 & 182 \\
\textbf{nutrition} & 0.66 & 0.53 & 0.44 & 0.49 & 0.63 & 294 \\
\textbf{management} & 0.72 & 0.59 & 0.59 & 0.63 & 0.76 & 99 \\
\textbf{jurisprudence} & 0.63 & 0.43 & 0.39 & 0.49 & 0.66 & 103 \\
\textbf{human aging} & 0.60 & 0.44 & 0.38 & 0.46 & 0.56 & 211 \\
\textbf{college chemistry} & 0.55 & 0.51 & 0.38 & 0.43 & 0.45 & 88 \\
\textbf{business ethics} & 0.45 & 0.52 & 0.43 & 0.42 & 0.51 & 88 \\
\textbf{high school psychology} & 0.67 & 0.56 & 0.56 & 0.61 & 0.71 & 513 \\
\textbf{conceptual physics} & 0.59 & 0.51 & 0.38 & 0.36 & 0.40 & 230 \\
\textbf{prehistory} & 0.68 & 0.57 & 0.44 & 0.54 & 0.61 & 297 \\
\textbf{high school chemistry} & 0.54 & 0.47 & 0.32 & 0.37 & 0.46 & 191 \\
\textbf{high school world history} & 0.67 & 0.51 & 0.42 & 0.43 & 0.70 & 191 \\
\textbf{college biology} & 0.66 & 0.48 & 0.44 & 0.48 & 0.48 & 130 \\
\textbf{high school physics} & 0.49 & 0.41 & 0.34 & 0.34 & 0.30 & 146 \\
\textbf{high school european history} & 0.62 & 0.50 & 0.50 & 0.56 & 0.64 & 135 \\
\textbf{college computer science} & 0.52 & 0.42 & 0.27 & 0.38 & 0.36 & 96 \\
\textbf{us foreign policy} & 0.69 & 0.66 & 0.57 & 0.67 & 0.81 & 96 \\
\textbf{moral disputes} & 0.62 & 0.48 & 0.33 & 0.42 & 0.54 & 328 \\
\textbf{world religions} & 0.69 & 0.58 & 0.55 & 0.62 & 0.75 & 163 \\
\textbf{high school statistics} & 0.55 & 0.43 & 0.40 & 0.47 & 0.44 & 199 \\
\textbf{international law} & 0.52 & 0.48 & 0.48 & 0.48 & 0.71 & 108 \\
\textbf{security studies} & 0.84 & 0.58 & 0.41 & 0.49 & 0.64 & 222 \\
\textbf{professional medicine} & 0.59 & 0.42 & 0.52 & 0.53 & 0.53 & 257 \\
\textbf{marketing} & 0.74 & 0.63 & 0.56 & 0.65 & 0.77 & 226 \\
\textbf{high school us history} & 0.61 & 0.53 & 0.45 & 0.49 & 0.66 & 179 \\
\textbf{sociology} & 0.77 & 0.57 & 0.52 & 0.60 & 0.75 & 190 \\
\textbf{anatomy} & 0.66 & 0.47 & 0.37 & 0.45 & 0.49 & 133 \\
\textbf{virology} & 0.63 & 0.61 & 0.39 & 0.41 & 0.43 & 147 \\
\textbf{professional psychology} & 0.63 & 0.48 & 0.39 & 0.45 & 0.53 & 575 \\
\textbf{miscellaneous} & 0.69 & 0.61 & 0.58 & 0.59 & 0.73 & 752 \\
\textbf{high school microeconomics} & 0.72 & 0.43 & 0.45 & 0.48 & 0.53 & 220 \\
\textbf{global facts} & 0.30 & 0.42 & 0.37 & 0.23 & 0.32 & 99 \\
\textbf{philosophy} & 0.72 & 0.51 & 0.42 & 0.48 & 0.65 & 296 \\
\textbf{college medicine} & 0.62 & 0.51 & 0.46 & 0.48 & 0.51 & 162 \\
\textbf{professional accounting} & 0.59 & 0.30 & 0.32 & 0.36 & 0.40 & 266 \\
\hline
\end{tabular}
}
\caption{Detailed results of LLaMA-13B on different categories of MMLU.}
\label{tab:models_agreement_llama_13b}
\end{table*}

\begin{table*}[!]
\centering
\resizebox{\linewidth}{!}{
\begin{tabular}{|l|r|r|r|r|r|r|}
\hline
\textbf{Category} & \textbf{Agreement(Label)} & \textbf{Agreement(Seq)} & \textbf{Acc.(Gen)} & \textbf{Acc.(Label)} & \textbf{Acc.(Seq)} & \textbf{Examples} \\ \hline
\textbf{moral scenarios} & 0.29 & 0.47 & 0.32 & 0.24 & 0.27 & 893 \\
\textbf{college physics} & 0.74 & 0.57 & 0.24 & 0.27 & 0.27 & 100 \\
\textbf{high school biology} & 0.83 & 0.69 & 0.58 & 0.58 & 0.64 & 309 \\
\textbf{college mathematics} & 0.89 & 0.71 & 0.26 & 0.29 & 0.29 & 100 \\
\textbf{abstract algebra} & 0.41 & 0.63 & 0.34 & 0.26 & 0.29 & 99 \\
\textbf{high school computer science} & 0.82 & 0.64 & 0.48 & 0.47 & 0.55 & 99 \\
\textbf{astronomy} & 0.83 & 0.64 & 0.53 & 0.57 & 0.58 & 152 \\
\textbf{computer security} & 0.76 & 0.61 & 0.57 & 0.60 & 0.66 & 100 \\
\textbf{logical fallacies} & 0.68 & 0.65 & 0.56 & 0.59 & 0.69 & 162 \\
\textbf{professional law} & 0.81 & 0.72 & 0.37 & 0.39 & 0.40 & 1500 \\
\textbf{clinical knowledge} & 0.78 & 0.70 & 0.55 & 0.54 & 0.59 & 262 \\
\textbf{elementary mathematics} & 0.72 & 0.60 & 0.33 & 0.30 & 0.32 & 374 \\
\textbf{high school macroeconomics} & 0.82 & 0.73 & 0.44 & 0.46 & 0.50 & 389 \\
\textbf{formal logic} & 0.63 & 0.48 & 0.24 & 0.30 & 0.24 & 122 \\
\textbf{high school government and politics} & 0.90 & 0.75 & 0.63 & 0.65 & 0.76 & 193 \\
\textbf{medical genetics} & 0.72 & 0.63 & 0.47 & 0.54 & 0.58 & 100 \\
\textbf{electrical engineering} & 0.74 & 0.68 & 0.50 & 0.51 & 0.54 & 145 \\
\textbf{high school mathematics} & 0.74 & 0.59 & 0.27 & 0.24 & 0.27 & 266 \\
\textbf{public relations} & 0.79 & 0.69 & 0.53 & 0.54 & 0.63 & 110 \\
\textbf{econometrics} & 0.78 & 0.70 & 0.26 & 0.31 & 0.24 & 111 \\
\textbf{machine learning} & 0.58 & 0.74 & 0.32 & 0.42 & 0.33 & 111 \\
\textbf{human sexuality} & 0.85 & 0.73 & 0.55 & 0.57 & 0.64 & 131 \\
\textbf{high school geography} & 0.85 & 0.69 & 0.59 & 0.60 & 0.65 & 198 \\
\textbf{nutrition} & 0.81 & 0.65 & 0.51 & 0.52 & 0.61 & 305 \\
\textbf{management} & 0.79 & 0.71 & 0.57 & 0.63 & 0.69 & 103 \\
\textbf{jurisprudence} & 0.72 & 0.58 & 0.51 & 0.60 & 0.69 & 108 \\
\textbf{human aging} & 0.80 & 0.66 & 0.45 & 0.53 & 0.62 & 221 \\
\textbf{college chemistry} & 0.78 & 0.65 & 0.28 & 0.35 & 0.34 & 95 \\
\textbf{business ethics} & 0.72 & 0.68 & 0.49 & 0.52 & 0.54 & 100 \\
\textbf{high school psychology} & 0.84 & 0.76 & 0.63 & 0.65 & 0.72 & 542 \\
\textbf{conceptual physics} & 0.83 & 0.64 & 0.36 & 0.37 & 0.41 & 235 \\
\textbf{prehistory} & 0.82 & 0.71 & 0.52 & 0.53 & 0.63 & 323 \\
\textbf{high school chemistry} & 0.73 & 0.63 & 0.38 & 0.38 & 0.43 & 203 \\
\textbf{high school world history} & 0.71 & 0.72 & 0.61 & 0.68 & 0.75 & 218 \\
\textbf{college biology} & 0.81 & 0.65 & 0.44 & 0.47 & 0.58 & 144 \\
\textbf{high school physics} & 0.79 & 0.55 & 0.36 & 0.35 & 0.33 & 148 \\
\textbf{high school european history} & 0.83 & 0.69 & 0.55 & 0.63 & 0.67 & 144 \\
\textbf{college computer science} & 0.86 & 0.70 & 0.37 & 0.33 & 0.43 & 99 \\
\textbf{us foreign policy} & 0.88 & 0.83 & 0.71 & 0.73 & 0.81 & 100 \\
\textbf{moral disputes} & 0.84 & 0.70 & 0.48 & 0.52 & 0.60 & 345 \\
\textbf{world religions} & 0.87 & 0.77 & 0.69 & 0.70 & 0.77 & 171 \\
\textbf{high school statistics} & 0.79 & 0.60 & 0.35 & 0.34 & 0.34 & 216 \\
\textbf{international law} & 0.78 & 0.71 & 0.61 & 0.68 & 0.72 & 120 \\
\textbf{security studies} & 0.87 & 0.68 & 0.52 & 0.55 & 0.66 & 241 \\
\textbf{professional medicine} & 0.66 & 0.63 & 0.46 & 0.42 & 0.50 & 265 \\
\textbf{marketing} & 0.88 & 0.75 & 0.69 & 0.70 & 0.80 & 234 \\
\textbf{high school us history} & 0.71 & 0.69 & 0.58 & 0.64 & 0.74 & 200 \\
\textbf{sociology} & 0.86 & 0.73 & 0.65 & 0.71 & 0.75 & 201 \\
\textbf{anatomy} & 0.82 & 0.73 & 0.47 & 0.46 & 0.52 & 135 \\
\textbf{virology} & 0.74 & 0.62 & 0.37 & 0.44 & 0.47 & 165 \\
\textbf{professional psychology} & 0.78 & 0.68 & 0.47 & 0.51 & 0.54 & 610 \\
\textbf{miscellaneous} & 0.82 & 0.72 & 0.66 & 0.69 & 0.77 & 782 \\
\textbf{high school microeconomics} & 0.74 & 0.62 & 0.46 & 0.45 & 0.51 & 238 \\
\textbf{global facts} & 0.80 & 0.66 & 0.32 & 0.31 & 0.31 & 100 \\
\textbf{philosophy} & 0.83 & 0.72 & 0.55 & 0.55 & 0.65 & 310 \\
\textbf{college medicine} & 0.80 & 0.63 & 0.41 & 0.43 & 0.42 & 167 \\
\textbf{professional accounting} & 0.80 & 0.66 & 0.37 & 0.39 & 0.41 & 282 \\
\hline
\end{tabular}
}
\caption{Detailed results of LLaMA-13B-chat on different categories of MMLU.}
\label{tab:models_agreement_llama_13b_chat}
\end{table*}

\begin{table*}[!]
\centering
\resizebox{\linewidth}{!}{
\begin{tabular}{|l|r|r|r|r|r|r|}
\hline
\textbf{Category} & \textbf{Agreement(Label)} & \textbf{Agreement(Seq)} & \textbf{Acc.(Gen)} & \textbf{Acc.(Label)} & \textbf{Acc.(Seq)} & \textbf{Examples} \\ \hline
\textbf{moral scenarios} & 0.64 & 0.98 & 0.24 & 0.25 & 0.24 & 878 \\
\textbf{college physics} & 0.31 & 0.50 & 0.31 & 0.21 & 0.44 & 96 \\
\textbf{high school biology} & 0.44 & 0.68 & 0.65 & 0.47 & 0.73 & 303 \\
\textbf{college mathematics} & 0.31 & 0.48 & 0.24 & 0.35 & 0.34 & 94 \\
\textbf{abstract algebra} & 0.26 & 0.48 & 0.40 & 0.19 & 0.30 & 96 \\
\textbf{high school computer science} & 0.41 & 0.55 & 0.53 & 0.47 & 0.64 & 92 \\
\textbf{astronomy} & 0.41 & 0.57 & 0.59 & 0.39 & 0.61 & 148 \\
\textbf{computer security} & 0.49 & 0.70 & 0.61 & 0.49 & 0.74 & 92 \\
\textbf{logical fallacies} & 0.50 & 0.70 & 0.66 & 0.48 & 0.75 & 159 \\
\textbf{professional law} & 0.36 & 0.58 & 0.39 & 0.30 & 0.44 & 1508 \\
\textbf{clinical knowledge} & 0.47 & 0.66 & 0.63 & 0.44 & 0.69 & 261 \\
\textbf{elementary mathematics} & 0.32 & 0.51 & 0.43 & 0.29 & 0.40 & 373 \\
\textbf{high school macroeconomics} & 0.37 & 0.59 & 0.51 & 0.35 & 0.59 & 384 \\
\textbf{formal logic} & 0.44 & 0.53 & 0.36 & 0.24 & 0.35 & 110 \\
\textbf{high school government and politics} & 0.50 & 0.71 & 0.74 & 0.53 & 0.84 & 191 \\
\textbf{medical genetics} & 0.52 & 0.61 & 0.61 & 0.52 & 0.69 & 100 \\
\textbf{electrical engineering} & 0.42 & 0.62 & 0.50 & 0.40 & 0.58 & 141 \\
\textbf{high school mathematics} & 0.30 & 0.44 & 0.34 & 0.27 & 0.35 & 250 \\
\textbf{public relations} & 0.54 & 0.60 & 0.58 & 0.42 & 0.66 & 106 \\
\textbf{econometrics} & 0.43 & 0.61 & 0.41 & 0.28 & 0.44 & 113 \\
\textbf{machine learning} & 0.26 & 0.37 & 0.38 & 0.31 & 0.48 & 108 \\
\textbf{human sexuality} & 0.45 & 0.64 & 0.62 & 0.47 & 0.75 & 130 \\
\textbf{high school geography} & 0.58 & 0.70 & 0.66 & 0.51 & 0.75 & 188 \\
\textbf{nutrition} & 0.46 & 0.63 & 0.60 & 0.46 & 0.70 & 301 \\
\textbf{management} & 0.55 & 0.70 & 0.66 & 0.43 & 0.80 & 100 \\
\textbf{jurisprudence} & 0.41 & 0.62 & 0.51 & 0.38 & 0.74 & 104 \\
\textbf{human aging} & 0.39 & 0.59 & 0.56 & 0.49 & 0.66 & 216 \\
\textbf{college chemistry} & 0.33 & 0.39 & 0.30 & 0.28 & 0.47 & 99 \\
\textbf{business ethics} & 0.32 & 0.60 & 0.53 & 0.35 & 0.58 & 96 \\
\textbf{high school psychology} & 0.52 & 0.75 & 0.73 & 0.48 & 0.78 & 530 \\
\textbf{conceptual physics} & 0.43 & 0.57 & 0.50 & 0.39 & 0.53 & 230 \\
\textbf{prehistory} & 0.43 & 0.71 & 0.59 & 0.39 & 0.71 & 318 \\
\textbf{high school chemistry} & 0.32 & 0.59 & 0.43 & 0.29 & 0.50 & 197 \\
\textbf{high school world history} & 0.33 & 0.59 & 0.63 & 0.46 & 0.79 & 212 \\
\textbf{college biology} & 0.41 & 0.67 & 0.57 & 0.41 & 0.67 & 141 \\
\textbf{high school physics} & 0.33 & 0.46 & 0.34 & 0.27 & 0.30 & 146 \\
\textbf{high school european history} & 0.33 & 0.69 & 0.57 & 0.36 & 0.77 & 143 \\
\textbf{college computer science} & 0.29 & 0.47 & 0.33 & 0.32 & 0.54 & 96 \\
\textbf{us foreign policy} & 0.59 & 0.79 & 0.78 & 0.60 & 0.84 & 100 \\
\textbf{moral disputes} & 0.41 & 0.64 & 0.56 & 0.38 & 0.68 & 338 \\
\textbf{world religions} & 0.52 & 0.81 & 0.75 & 0.53 & 0.81 & 165 \\
\textbf{high school statistics} & 0.35 & 0.55 & 0.38 & 0.30 & 0.46 & 207 \\
\textbf{international law} & 0.45 & 0.66 & 0.61 & 0.47 & 0.76 & 119 \\
\textbf{security studies} & 0.40 & 0.62 & 0.56 & 0.39 & 0.70 & 241 \\
\textbf{professional medicine} & 0.42 & 0.63 & 0.56 & 0.42 & 0.68 & 268 \\
\textbf{marketing} & 0.58 & 0.77 & 0.81 & 0.59 & 0.86 & 226 \\
\textbf{high school us history} & 0.34 & 0.63 & 0.63 & 0.39 & 0.76 & 197 \\
\textbf{sociology} & 0.49 & 0.79 & 0.69 & 0.54 & 0.86 & 200 \\
\textbf{anatomy} & 0.44 & 0.62 & 0.54 & 0.32 & 0.56 & 133 \\
\textbf{virology} & 0.47 & 0.66 & 0.51 & 0.34 & 0.52 & 161 \\
\textbf{professional psychology} & 0.48 & 0.65 & 0.56 & 0.39 & 0.61 & 604 \\
\textbf{miscellaneous} & 0.53 & 0.73 & 0.72 & 0.53 & 0.79 & 769 \\
\textbf{high school microeconomics} & 0.38 & 0.61 & 0.56 & 0.36 & 0.63 & 233 \\
\textbf{global facts} & 0.42 & 0.50 & 0.43 & 0.26 & 0.41 & 92 \\
\textbf{philosophy} & 0.43 & 0.67 & 0.61 & 0.37 & 0.69 & 289 \\
\textbf{college medicine} & 0.40 & 0.64 & 0.54 & 0.33 & 0.60 & 164 \\
\textbf{professional accounting} & 0.38 & 0.53 & 0.46 & 0.34 & 0.47 & 268 \\
\hline
\end{tabular}
}
\caption{Detailed results of Mistral-7B on different categories of MMLU.}
\label{tab:models_agreement_mistral}
\end{table*}

\begin{table*}[!]
\centering
\resizebox{\linewidth}{!}{
\begin{tabular}{|l|r|r|r|r|r|r|}
\hline
\textbf{Category} & \textbf{Agreement(Label)} & \textbf{Agreement(Seq)} & \textbf{Acc.(Gen)} & \textbf{Acc.(Label)} & \textbf{Acc.(Seq)} & \textbf{Examples} \\ \hline
\textbf{moral scenarios} & 0.08 & 0.02 & 0.28 & 0.23 & 0.24 & 894 \\
\textbf{college physics} & 0.34 & 0.48 & 0.26 & 0.16 & 0.29 & 100 \\
\textbf{high school biology} & 0.43 & 0.64 & 0.57 & 0.39 & 0.65 & 310 \\
\textbf{college mathematics} & 0.21 & 0.37 & 0.29 & 0.24 & 0.39 & 97 \\
\textbf{abstract algebra} & 0.11 & 0.27 & 0.34 & 0.17 & 0.33 & 99 \\
\textbf{high school computer science} & 0.36 & 0.60 & 0.51 & 0.42 & 0.50 & 100 \\
\textbf{astronomy} & 0.41 & 0.57 & 0.52 & 0.34 & 0.53 & 152 \\
\textbf{computer security} & 0.42 & 0.56 & 0.52 & 0.52 & 0.65 & 100 \\
\textbf{logical fallacies} & 0.50 & 0.69 & 0.59 & 0.47 & 0.71 & 163 \\
\textbf{professional law} & 0.37 & 0.56 & 0.34 & 0.30 & 0.40 & 1521 \\
\textbf{clinical knowledge} & 0.39 & 0.66 & 0.56 & 0.41 & 0.61 & 265 \\
\textbf{elementary mathematics} & 0.32 & 0.49 & 0.45 & 0.26 & 0.34 & 374 \\
\textbf{high school macroeconomics} & 0.35 & 0.56 & 0.44 & 0.28 & 0.51 & 389 \\
\textbf{formal logic} & 0.23 & 0.39 & 0.36 & 0.30 & 0.38 & 122 \\
\textbf{high school government and politics} & 0.51 & 0.68 & 0.60 & 0.44 & 0.72 & 193 \\
\textbf{medical genetics} & 0.46 & 0.59 & 0.52 & 0.51 & 0.63 & 100 \\
\textbf{electrical engineering} & 0.38 & 0.57 & 0.50 & 0.37 & 0.54 & 143 \\
\textbf{high school mathematics} & 0.36 & 0.38 & 0.27 & 0.22 & 0.30 & 256 \\
\textbf{public relations} & 0.49 & 0.73 & 0.51 & 0.34 & 0.57 & 110 \\
\textbf{econometrics} & 0.39 & 0.49 & 0.30 & 0.28 & 0.32 & 114 \\
\textbf{machine learning} & 0.21 & 0.30 & 0.29 & 0.33 & 0.46 & 112 \\
\textbf{human sexuality} & 0.47 & 0.64 & 0.56 & 0.46 & 0.69 & 129 \\
\textbf{high school geography} & 0.53 & 0.68 & 0.57 & 0.47 & 0.67 & 198 \\
\textbf{nutrition} & 0.43 & 0.59 & 0.49 & 0.40 & 0.63 & 306 \\
\textbf{management} & 0.51 & 0.68 & 0.60 & 0.47 & 0.74 & 103 \\
\textbf{jurisprudence} & 0.44 & 0.61 & 0.52 & 0.42 & 0.67 & 108 \\
\textbf{human aging} & 0.46 & 0.61 & 0.51 & 0.48 & 0.60 & 223 \\
\textbf{college chemistry} & 0.36 & 0.44 & 0.32 & 0.29 & 0.37 & 97 \\
\textbf{business ethics} & 0.40 & 0.52 & 0.52 & 0.39 & 0.58 & 100 \\
\textbf{high school psychology} & 0.51 & 0.71 & 0.65 & 0.47 & 0.72 & 545 \\
\textbf{conceptual physics} & 0.40 & 0.53 & 0.43 & 0.31 & 0.46 & 235 \\
\textbf{prehistory} & 0.40 & 0.66 & 0.54 & 0.39 & 0.58 & 323 \\
\textbf{high school chemistry} & 0.32 & 0.47 & 0.41 & 0.24 & 0.43 & 200 \\
\textbf{high school world history} & 0.40 & 0.62 & 0.57 & 0.47 & 0.75 & 223 \\
\textbf{college biology} & 0.40 & 0.60 & 0.51 & 0.35 & 0.60 & 144 \\
\textbf{high school physics} & 0.32 & 0.57 & 0.25 & 0.23 & 0.32 & 146 \\
\textbf{high school european history} & 0.37 & 0.67 & 0.56 & 0.33 & 0.67 & 147 \\
\textbf{college computer science} & 0.32 & 0.50 & 0.30 & 0.30 & 0.46 & 96 \\
\textbf{us foreign policy} & 0.56 & 0.75 & 0.63 & 0.57 & 0.76 & 100 \\
\textbf{moral disputes} & 0.47 & 0.66 & 0.53 & 0.40 & 0.59 & 345 \\
\textbf{world religions} & 0.52 & 0.67 & 0.59 & 0.52 & 0.69 & 171 \\
\textbf{high school statistics} & 0.38 & 0.51 & 0.38 & 0.25 & 0.41 & 213 \\
\textbf{international law} & 0.42 & 0.74 & 0.64 & 0.43 & 0.70 & 121 \\
\textbf{security studies} & 0.38 & 0.56 & 0.45 & 0.39 & 0.66 & 244 \\
\textbf{professional medicine} & 0.38 & 0.57 & 0.46 & 0.35 & 0.59 & 268 \\
\textbf{marketing} & 0.56 & 0.72 & 0.73 & 0.60 & 0.80 & 234 \\
\textbf{high school us history} & 0.40 & 0.59 & 0.52 & 0.41 & 0.72 & 202 \\
\textbf{sociology} & 0.52 & 0.76 & 0.67 & 0.52 & 0.78 & 201 \\
\textbf{anatomy} & 0.31 & 0.57 & 0.48 & 0.25 & 0.47 & 135 \\
\textbf{virology} & 0.39 & 0.58 & 0.36 & 0.33 & 0.42 & 166 \\
\textbf{professional psychology} & 0.46 & 0.62 & 0.48 & 0.37 & 0.50 & 611 \\
\textbf{miscellaneous} & 0.51 & 0.72 & 0.69 & 0.51 & 0.75 & 783 \\
\textbf{high school microeconomics} & 0.36 & 0.58 & 0.50 & 0.37 & 0.60 & 237 \\
\textbf{global facts} & 0.40 & 0.54 & 0.36 & 0.23 & 0.31 & 100 \\
\textbf{philosophy} & 0.49 & 0.66 & 0.52 & 0.36 & 0.60 & 311 \\
\textbf{college medicine} & 0.40 & 0.61 & 0.40 & 0.27 & 0.53 & 168 \\
\textbf{professional accounting} & 0.39 & 0.54 & 0.36 & 0.29 & 0.39 & 282 \\
\hline
\end{tabular}
}
\caption{Detailed results of Mistral-7B-Chat on different categories of MMLU.}
\label{tab:models_agreement_mistral_chat}
\end{table*}

\begin{table*}[ht!]
    \centering
    \resizebox{0.75\linewidth}{!}{
        \begin{tabular}{lcccccc}
            \toprule
            & \multicolumn{2}{c}{MMLU} & \multicolumn{2}{c}{Truthful-QA} & \multicolumn{2}{c}{Belebele} \\
            \cmidrule(lr){2-3} \cmidrule(lr){4-5} \cmidrule(lr){6-7}
            Model & Label-Gen & Seq-Gen & Label-Gen & Seq-Gen & Label-Gen & Seq-Gen \\
            \midrule
            Mistral-7B & -9.3 & 12.1 & -3.7 & -14.1 & 9.3 & -3.9 \\
            Mistral-7B-Instruct & -8.0 & 8.9 & 14.7 & 7.5 & 6.7 & 4.5 \\
            LLaMA-1-7B & -12.1 & -13.4 & 29.7 & 11.5 & 24.3 & -4.6 \\
            Vicuna-7B & 3.9 & 9.8 & 27.8 & 12.1 & 4.0 & 16.5 \\
            LLaMA-2-7B & 36.7 & -4.1 & 5.1 & 2.6 & 3.3 & -6.4 \\
            LLaMA-2-7B-chat & 41.4 & 13.9 & 22.4 & -33.7 & 4.8 & 1.1 \\
            LLaMA-2-13B & 17.4 & 7.8 & 9.2 & -22.7 & 6.6 & 2.8 \\
            LLaMA-2-13B-chat & 29.2 & 20.0 & 25.4 & -12.3 & 7.9 & 7.3 \\
            LLaMA-2-70B & 15.4 & 4.6 & 7.3 & -24.8 & 1.2 & -2.0 \\
            LLaMA-2-70B-chat & 29.0 & 16.2 & 22.5 & -20.8 & 2.6 & 1.0 \\
            \bottomrule
        \end{tabular}
    }
    \caption{Differences in label and sequence accuracies compared to generation accuracies across datasets.}
    \label{tab:accuracy_differences}
\end{table*}

\subsection{Experimental Setup}
\subsubsection{Datasets}

\paragraph{MMLU} The Massive Multitask Language Understanding~(MMLU)~\cite{DBLP:conf/iclr/HendrycksBBZMSS21_mmlu} benchmark is a comprehensive test designed to assess knowledge acquired during pretraining of language models, especially in zero-shot and few-shot settings. Introduced by~\citep{DBLP:conf/iclr/HendrycksBBZMSS21_mmlu}., MMLU encompasses 57 subjects across diverse fields including STEM, humanities, social sciences, and others, making it a broad measure of both world knowledge and problem-solving ability~\cite{DBLP:conf/iclr/HendrycksBBZMSS21_mmlu}. The dataset contains 17,803 examples with a range of difficulties, from elementary to advanced professional levels. Its comprehensive nature allows for a detailed examination of a model’s strengths and weaknesses across various disciplines.

\paragraph{Truthful-QA} The Truthful-QA dataset~\cite{lin-etal-2022-truthfulqa} is a benchmark to assess the truthfulness of language model responses to questions. This dataset contains 817 questions spanning 38 diverse categories, including health, law, finance, and politics. The key characteristic of Truthful-QA is its design to elicit imitative falsehoods, wherein some questions are crafted to provoke false answers based on common misconceptions or false beliefs. The dataset aims to test language models' ability to avoid generating false answers that may have been learned through imitating human texts. Importantly, the Truthful-QA questions are adversarial in nature, designed to pinpoint weaknesses in the truthfulness of language models. Additionally, it features a set of true and false reference answers for each question, backed by reliable sources. 


\paragraph{Belebele} The Belebele Benchmark~\cite{bandarkar2023belebele} is a massively multilingual reading comprehension dataset designed to evaluate machine reading comprehension (MRC) capabilities across various languages. Developed by Facebook Research, it features 900 multiple-choice questions per language, spanning 122 language variants, totaling 109,800 questions linked to 488 distinct passages. Each question has four answer options, with only one correct answer. This benchmark encompasses a wide range of languages, from high-resource to low-resource, making it ideal for assessing the performance of language models in diverse linguistic contexts.

\subsubsection{Models}

\paragraph{LLaMA}
LLaMA-1~\cite{touvron2023llama1}, Vicuna~\cite{vicuna2023} and LLaMA-2~\cite{touvron2023llama2} is a family of large language models (LLMs), encompassing a range of pretrained and fine-tuned generative text models with parameters varying from 7 billion to 70 billion. The model was trained on a new mix of publicly available online data, with a considerable size of 2 trillion tokens, and includes over one million human-annotated examples for fine-tuning. Its training and evaluation emphasize both performance and safety. These fine-tuned models have shown superior performance in human evaluations for helpfulness and safety, matching or even surpassing other well-known models like ChatGPT and PaLM in certain aspects.

\paragraph{Mistral}
The Mistral model~\cite{jiang2023mistral} equipped with 7.3 billion parameters, is designed to outperform its counterparts in terms of efficiency and effectiveness. Notable features of Mistral 7B include its proficiency in outperforming LLaMA-2-13B~\cite{touvron2023llama2} across various benchmarks and approaching the performance of CodeLLaMA-7B~\cite{rozière2023code_llama} in code-related tasks while maintaining strong English language capabilities. Additionally, Mistral 7B incorporates Grouped-query attention (GQA) for faster inference and Sliding Window Attention (SWA) to manage longer sequences more economically.

\paragraph{lm-harness} The lm-harness ~\cite{eval-harness}~\footnote{\url{https://github.com/EleutherAI/lm-evaluation-harness}}, developed by EleutherAI, is a comprehensive framework designed for the few-shot evaluation of autoregressive language models. This library is pivotal in the field of natural language processing for assessing the performance of language models in few-shot settings. It stands out due to its versatility and ability to handle a variety of language models, making it a valuable tool for researchers in the field. The lm-harness library facilitates robust and efficient evaluations, contributing significantly to advancements in language model development and assessment~\cite{eval-harness}.

\subsection{Elo-based Chatbot Arena Leaderboard}
\label{sec:arena}
In the Elo-based Chatbot Arena Leaderboard, crowds are given an interface to ask questions to LLMs. The users are then given 2 options from 2 anonymous LLMs, in which the user has to vote for the better one, which will be the winner LLM. Based on several win-lose interactions, we can then calculate the Elo score.

Elo scores have been previously designed in rank multiple players that involve multiple matches across different people, such as chess. It is good for determining a unified ranking across every player (in this case, LLMs). From the Elo score of 2 players, we can predict the winning chance of both players. For example, an LLM with an Elo of 1200 will win against an LLM with an Elo of 900 85\% of the time.

Chatbot Arena is one of the popular Elo-based leaderboards. It supports a variety of LLMs, both proprietary and open-sourced, and has accumulated hundreds of thousands of votes.